\documentclass[10pt,twocolumn,letterpaper]{article}

\usepackage{hhline}
\usepackage{color, colortbl}
\usepackage{multirow}
\usepackage{wacv}
\usepackage{times}
\usepackage{epsfig}
\usepackage{graphicx}
\usepackage{amsmath}
\usepackage{amssymb}
\usepackage{subcaption}
\usepackage[utf8]{inputenc}
\usepackage{lipsum}  
\usepackage{booktabs}
\usepackage{xcolor}
\usepackage{makecell}
\usepackage{url}

\usepackage{authblk}
\usepackage{multirow}

\wacvfinalcopy 

\def\wacvPaperID{2} 

\ifwacvfinal\fi
\begin{document}

\title{Analysis of Gender Inequality In Face Recognition Accuracy}


\author[1]{Vítor Albiero}
\author[2]{Krishnapriya K.S.}
\author[2]{ Kushal Vangara}
\author[1]{\protect\\Kai Zhang}
\author[2]{Michael C. King}
\author[1]{Kevin W. Bowyer}
\affil[1]{University of Notre Dame, Notre Dame, Indiana}
\affil[2]{Florida Institute of Technology, Melbourne, Florida}

\maketitle
\thispagestyle{empty}

\begin{abstract}
We present a comprehensive analysis of how and why face recognition accuracy differs between men and women.
We show that accuracy is lower for women due to the combination of (1) the impostor distribution for women having a skew toward higher similarity scores, and (2) the genuine distribution for women having a skew toward lower similarity scores.
We show that this phenomenon of the impostor and genuine distributions for women shifting closer towards each other is general across datasets of African-American, Caucasian, and Asian faces. We show that the distribution of facial expressions may differ between male/female, but that the accuracy difference persists for image subsets rated confidently as neutral expression.
The accuracy difference also persists for image subsets rated as close to zero pitch angle.
Even when removing images with forehead partially occluded by hair/hat, the same impostor/genuine accuracy difference persists.
We show that the female genuine distribution improves when only female images without facial cosmetics are used, but that the female impostor distribution also degrades at the same time.
Lastly, we show that the accuracy difference persists even if a state-of-the-art deep learning method is trained from scratch using training data explicitly balanced between male and female images and subjects.
\end{abstract}

\section{Introduction}

Variation in face recognition accuracy based on gender, race or age has recently become a controversial topic \cite{Gtown, ACLU, MIT, Lohr2018}. Unequal accuracy across demographic groups can potentially undermine public acceptance of face recognition technology. 
Also, estimating accuracy based on images with a different demographic mix than the users of the technology can lead to unexpected problems in the operational scenario.
Therefore, it is important to understand what accuracy differences actually exist, and why.

This paper focuses on the issue of unequal face recognition accuracy between men and women. 
Previous work has concluded that face recognition is more accurate for men than for women. 
In order to better understand why the Receiver Operator Characteristic (ROC) curve for women is worse than the ROC for men, we examine differences in the underlying impostor and genuine distributions. 
Facial expression, head pose pitch, makeup use, and forehead occlusion by hair/hat have been speculated to cause differences in face recognition accuracy between men and women \cite{grother2010report, cook2018, Lu2018}, and so we report on experiments to determine if any of these factors can explain the observed differences in the impostor and genuine distributions.

We know of no previous work that reports on experiments designed to identify the cause of lower face recognition accuracy for women using deep learning methods.
As deep learning methods rely on training data, and training datasets typically have fewer images of women than men, we also report results for deep learning models trained on two perfectly gender balanced training datasets.


\begin{figure*}[t]
  \begin{subfigure}[b]{1\linewidth}
      \begin{subfigure}[b]{0.24\linewidth}
          \begin{subfigure}[b]{.49\columnwidth}
            \centering
            \includegraphics[width=\linewidth]{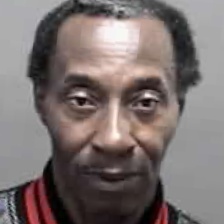}
          \end{subfigure}
          \hfill 
          \begin{subfigure}[b]{.49\columnwidth}
            \centering
            \includegraphics[width=\linewidth]{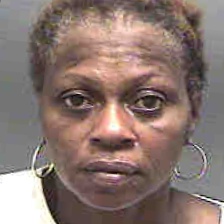}
          \end{subfigure}
          \caption{MORPH African American}
          \vspace{-0.5em}
      \end{subfigure}
      \hfill 
      \begin{subfigure}[b]{0.24\linewidth}
          \begin{subfigure}[b]{.49\columnwidth}
            \centering
            \includegraphics[width=\linewidth]{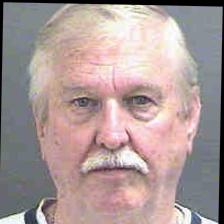}
          \end{subfigure}
          \hfill 
          \begin{subfigure}[b]{.49\columnwidth}
            \centering
            \includegraphics[width=\linewidth]{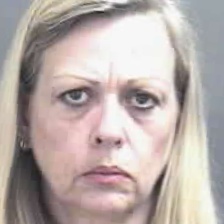}
          \end{subfigure}
          \caption{MORPH Caucasian}
          \vspace{-0.5em}
      \end{subfigure}
      \hfill 
      \begin{subfigure}[b]{0.24\linewidth}
          \begin{subfigure}[b]{.49\columnwidth}
            \centering
            \includegraphics[width=\linewidth]{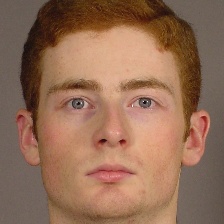}
          \end{subfigure}
          \hfill 
          \begin{subfigure}[b]{.49\columnwidth}
            \centering
            \includegraphics[width=\linewidth]{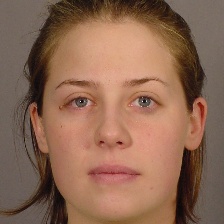}
          \end{subfigure}
          \caption{Notre Dame}
          \vspace{-0.5em}
      \end{subfigure}
      \hfill 
      \begin{subfigure}[b]{0.24\linewidth}
          \begin{subfigure}[b]{.49\columnwidth}
            \centering
            \includegraphics[width=\linewidth]{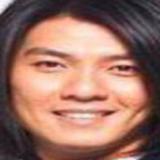}
          \end{subfigure}
          \hfill 
          \begin{subfigure}[b]{.49\columnwidth}
            \centering
            \includegraphics[width=\linewidth]{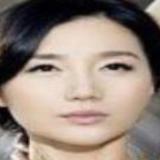}
          \end{subfigure}
          \caption{AFD}
          \vspace{-0.5em}
      \end{subfigure}
  \end{subfigure}
  \caption{Male and female samples from the datasets used.}
  \label{fig:mean_faces}
  \vspace{-1em}
\end{figure*}

\section{Literature Review}

To our knowledge, the earliest report that face recognition accuracy is lower for women was in the 2002 Face Recognition Vendor Test (FRVT) \cite{frvt}. In a 2009 meta-analysis of twenty-five prior works, Lui et al. \cite{Lui2009} found that men were slightly easier to recognize than women. They also noted that gender effects are often mixed with other covariates.

Beveridge et al. \cite{Beveridge2009} analyzed results for a Face Recognition Grand Challenge \cite{frgc} dataset that includes 159 women and 192 men. They reported that men had higher accuracy than women for two of the three algorithms, but women had higher accuracy for the third. 

Grother et al. \cite{Grother2010} analyzed the Multiple Biometric Evaluation (MBE) results and found that women have a higher false non-match rate (FNMR) at a fixed false match rate (FMR) for 6 of the 7 matchers evaluated. However, they did not investigate the relationship between impostor scores and gender. In terms of possible causes for unequal FNMR, one speculation offered is that women are generally shorter than men, and a camera adjusted for average male height may have non-optimal viewing angle for women.

In a 2012 paper, Klare et al. \cite{Klare2012} report on experiments with six matchers and a large dataset of mug-shot style images. They consider issues related to trainable and non-trainable algorithms, and balanced training data, and conclude that women are inherently more difficult to recognize. They also show results in which both the impostor and the genuine distributions for women are worse than for men.

Ueda et al. \cite{ueda2010influence} report that face recognition is advantaged by the presence of light makeup and disadvantaged by the presence of heavy makeup. 
They had face images of 24 Japanese women in their database with three makeup conditions – no makeup, light makeup, and heavy makeup. 
Their results show that the effect of makeup and gender on face recognition is statistically significant, with light makeup being the easiest to recognize, followed by no makeup and then heavy makeup. 

Guo et al. \cite{face_auth_makeup_changes} propose a correlation-based face recognition technique that is robust to changes introduced by facial makeup. 
These correlations are built on the extracted features from local patches on face images rather than the raw pixel values since the latter can be affected by facial makeup to some degree. 
Their patch selection scheme and discriminative mapping are found to be effective in improving facial makeup detection accuracy.

Dantcheva et al. \cite{can_facial_cosmetics} investigate how a technique called non-permanent facial makeup affects face recognition accuracy. 
They present results for the YouTube MakeUp (YMU) database, which has real makeup images, and the Virtual MakeUp (VMU) database, which has images modified with application of synthetic makeup. Results show a reduction in accuracy with all the matchers for images with makeup.


The works discussed above were performed prior to deep learning algorithms in face recognition. 
It is important to consider whether deep learning methods that achieve higher absolute accuracy show the same accuracy differences that were present in earlier algorithms.
In a presentation of results from the FRVT Ongoing effort at NIST, Grother \cite{Grother2017} notes that women are harder to recognize, as they have both FMR and FNMR higher than men.

Cook et al. \cite{cook2018} report a study in which images are acquired using eleven different automated systems and used with the same commercial matcher. 
They find large differences in genuine scores, and in acquisition times, across the 11 systems. 
Relative to this paper, they report that genuine scores were higher for men than women, suggesting that two images of the same female appear less alike on average than two images of the same male.
They speculate that possible causes could be hair-style and makeup.

Lu et al. \cite{Lu2018} reported a detailed study of the effect of various covariates, using deep CNN matchers and datasets from the IARPA Janus program. 
They report that accuracy for females is lower, and speculate that greater face occlusion by hairstyle and cosmetics use are possible causes.

In the 2019 Face Recognition Vendor Test \cite{frvt3} focused in demographic analysis, NIST reported that women have higher false positive rates than men (worst impostor distribution), and that the phenomenon is consistent across matchers and datasets.

Most previous works examining male/female difference in face recognition accuracy do not explicitly consider the differences in the impostor and genuine distributions that underlie the ROC curves. Also, while multiple previous works speculate that makeup could be the cause of lower accuracy for women \cite{Klare2012, cook2018, Lu2018}, after the deep learning wave, there has been no empirical investigation to test this speculation, or any of the other speculated causes such as pose angle,
facial expression, and occlusion by hair.

\section{Experimental Datasets and Matcher}

\begin{figure*}[t]
  \begin{subfigure}[b]{1\linewidth}
      \begin{subfigure}[b]{0.24\linewidth}
          \begin{subfigure}[b]{1\columnwidth}
            \centering
            \includegraphics[width=\linewidth]{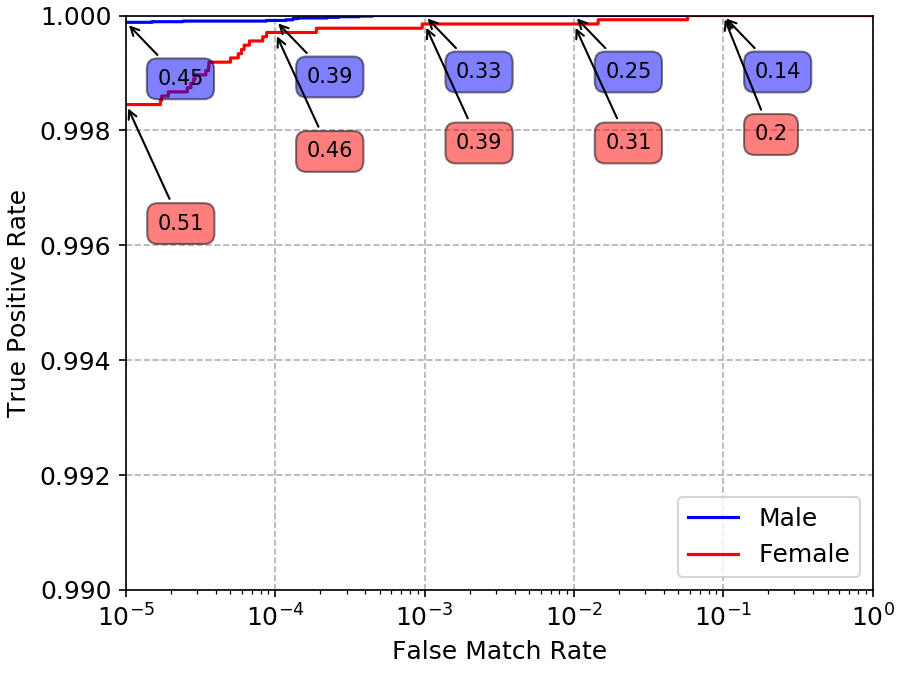}
          \end{subfigure}
          \caption{MORPH African American}
          \vspace{-0.5em}
      \end{subfigure}
      \hfill 
      \begin{subfigure}[b]{0.24\linewidth}
          \begin{subfigure}[b]{1\columnwidth}
            \centering
            \includegraphics[width=\linewidth]{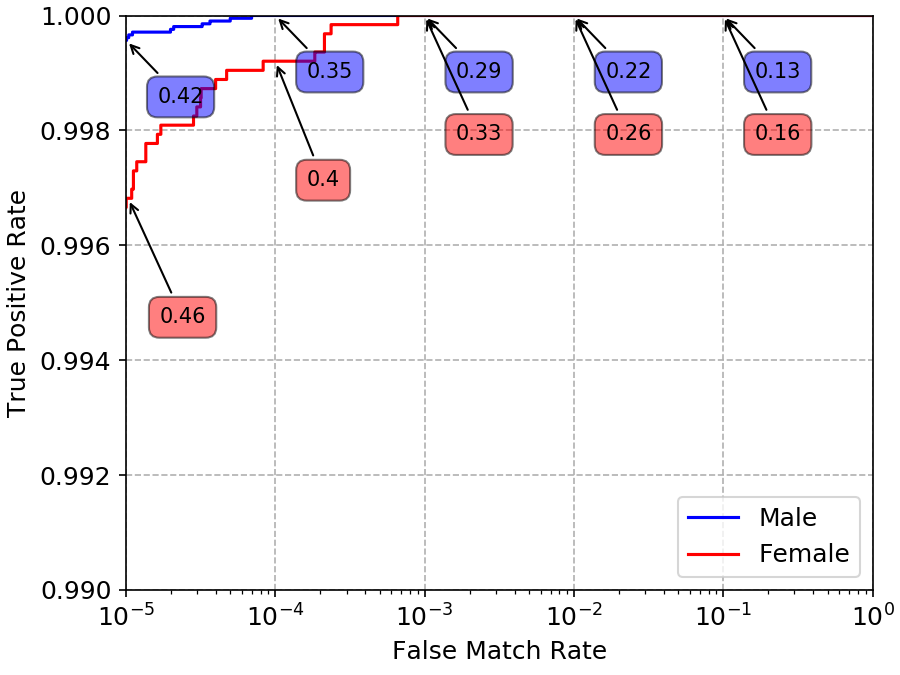}
          \end{subfigure}
          \caption{MORPH Caucasian}
          \vspace{-0.5em}
      \end{subfigure}
      \hfill 
      \begin{subfigure}[b]{0.24\linewidth}
          \begin{subfigure}[b]{1\columnwidth}
            \centering
            \includegraphics[width=\linewidth]{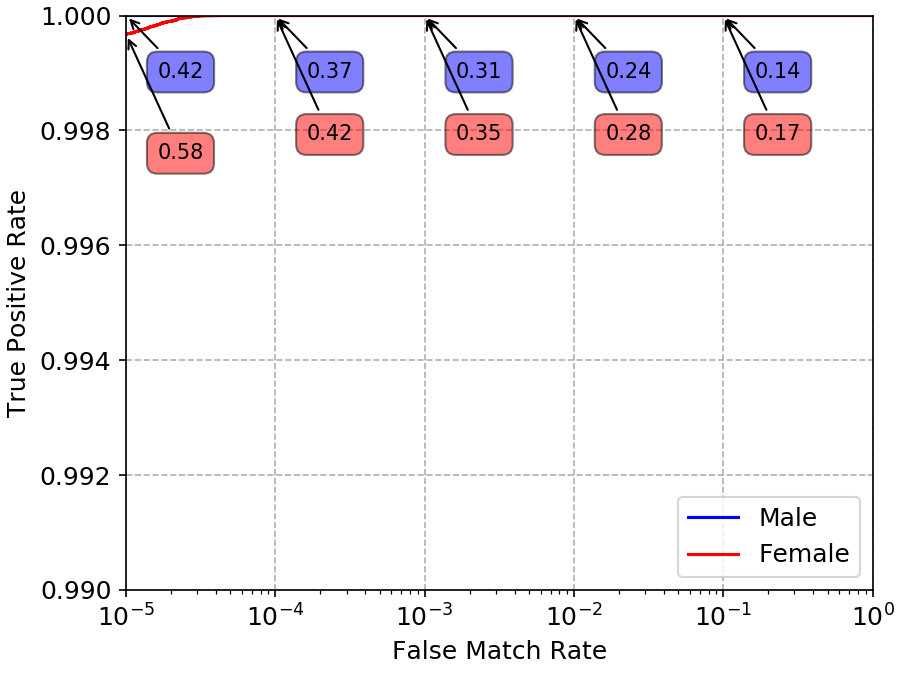}
          \end{subfigure}
          \caption{Notre Dame}
          \vspace{-0.5em}
      \end{subfigure}
      \hfill 
      \begin{subfigure}[b]{0.24\linewidth}
          \begin{subfigure}[b]{1\columnwidth}
            \centering
            \includegraphics[width=\linewidth]{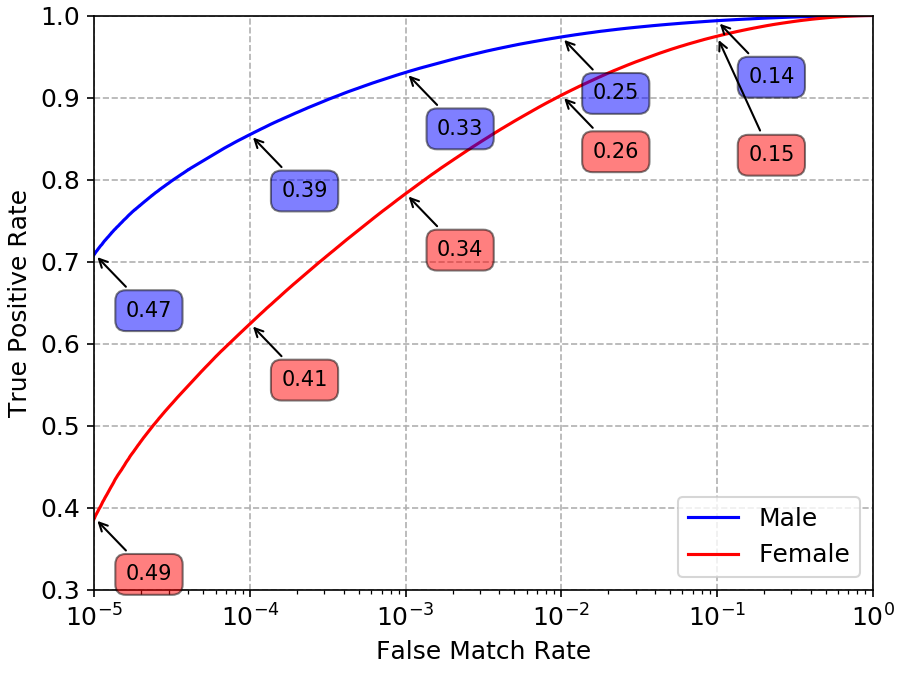}
          \end{subfigure}
          \caption{AFD}
          \vspace{-0.5em}
      \end{subfigure}
  \end{subfigure}
  \caption{Male and female ROC curves. Annotated values correspond to threshold used to achieve the specific FMR.}
  \label{fig:roc_original}
\end{figure*}

Male/female difference in face recognition accuracy is analyzed using four datasets.
Two of these datasets are the African-American and the Caucasian cohorts of the MORPH dataset \cite{morph}.
MORPH contains mugshot-style images of both African-Americans and Caucasians and was recently used in analysis of race-based differences in face recognition accuracy \cite{krishnapria_cvprw_2019, albiero2019does}.
Due to the accuracy differences based on race, the African-American and Caucasian cohorts are analyzed separately in this paper.
The African-American dataset contains 36,836 images of 8,864 males and 5,782 images of 1,500 females.
The Caucasian dataset contains 8,005 images of 2,132 males and 2,606 images of 637 females.
Subjects range in age approximately from 16 to 70 years old.

The Notre Dame dataset used in this work is drawn from image collections previously released by the University of Notre Dame \cite{frgc}.
It contains 14,345 images of 261 Caucasian males and 10,021 images of 169 Caucasian females.
Subjects are primarily younger.
All images have good quality and subjects appear in front of a uniform background.

The Asian Faces Dataset (AFD) \cite{afd} was assembled using ``in the wild'' images scraped from the web.
We curated a subset of AFD for this study by the following steps.
First, we identified a subset of mostly frontal face images.
Then we used a gender classifier on the frontal images, and manually checked any subject that had less than 75\% consistent gender prediction across their images.
Lastly, we removed mislabeled images, merged subjects that were the same person, and removed duplicate and near-duplicate images.
The curated version of AFD \footnote{\url{github.com/vitoralbiero/afd_dataset_cleaned}} used in our experiments contains 42,134 images of 911 males and 49,320 images of 967 females.

We report results for one of the best state-of-the-art open-source deep learning face matcher, ArcFace \cite{arcface}.
The ArcFace model \cite{insightface} used in this paper was trained with ResNet-100 using the MS1MV2 dataset, which is a curated version of the MS1M dataset \cite{ms1_celeb}.
We have experimented with other deep CNN matchers, and the pattern of results is similar, but the overall accuracy is usually noticeably lower.
For this reason, and to save space, we present only ArcFace results.
For MORPH and Notre Dame datasets, the faces are detected and aligned using RetinaFace \cite{retinaface}.
For AFD dataset, as the images were already tightly cropped, we do not perform alignment and use the images as they are.

\section{Experimental Results}
For all the experiments in this section, except for the AFD dataset (that does not have date of acquisition for its images), we only use non same-day images to generate the genuine pairs: subjects in MORPH have only one image captured per day; we explicitly did not match same-day genuine pairs in Notre Dame dataset.  

\begin{figure*}[t]
  \begin{subfigure}[b]{1\linewidth}
      \begin{subfigure}[b]{0.24\linewidth}
          \begin{subfigure}[b]{1\columnwidth}
            \centering
            \includegraphics[width=\linewidth]{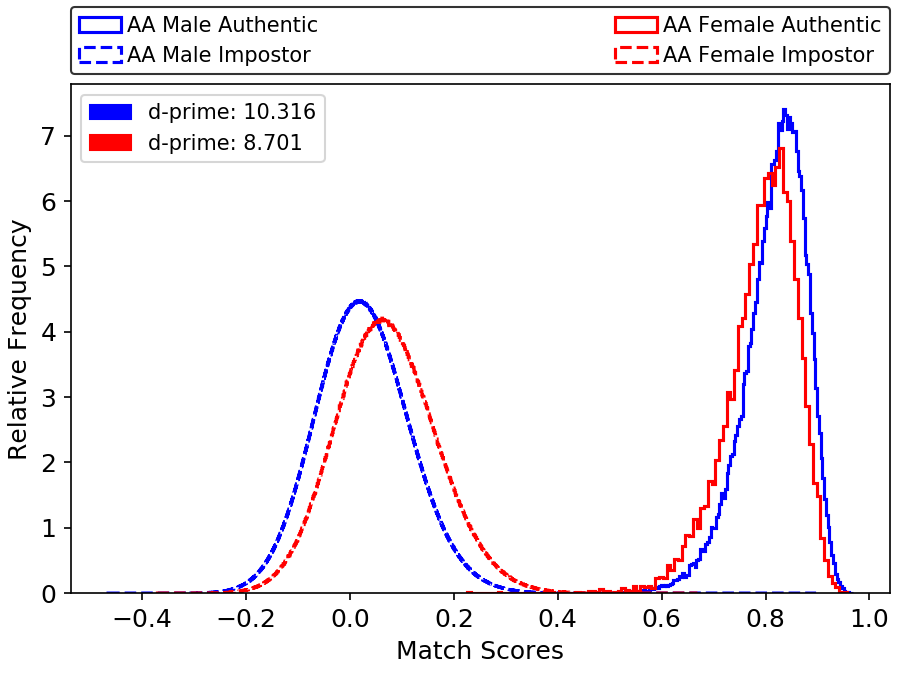}
          \end{subfigure}
          \caption{MORPH African American}
          \vspace{-0.5em}
      \end{subfigure}
      \hfill 
      \begin{subfigure}[b]{0.24\linewidth}
          \begin{subfigure}[b]{1\columnwidth}
            \centering
            \includegraphics[width=\linewidth]{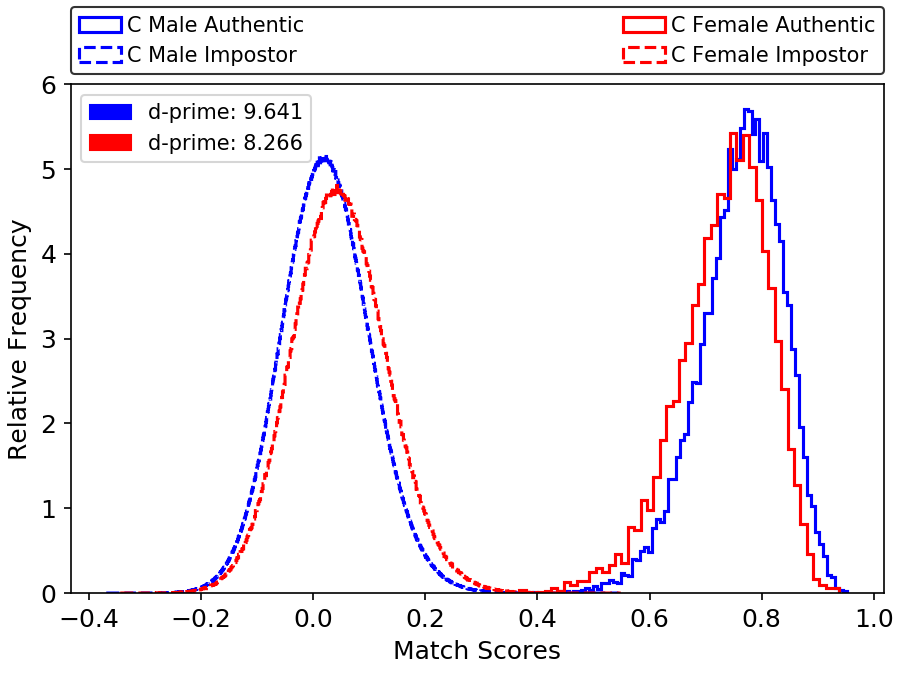}
          \end{subfigure}
          \caption{MORPH Caucasian}
          \vspace{-0.5em}
      \end{subfigure}
      \hfill 
      \begin{subfigure}[b]{0.24\linewidth}
          \begin{subfigure}[b]{1\columnwidth}
            \centering
            \includegraphics[width=\linewidth]{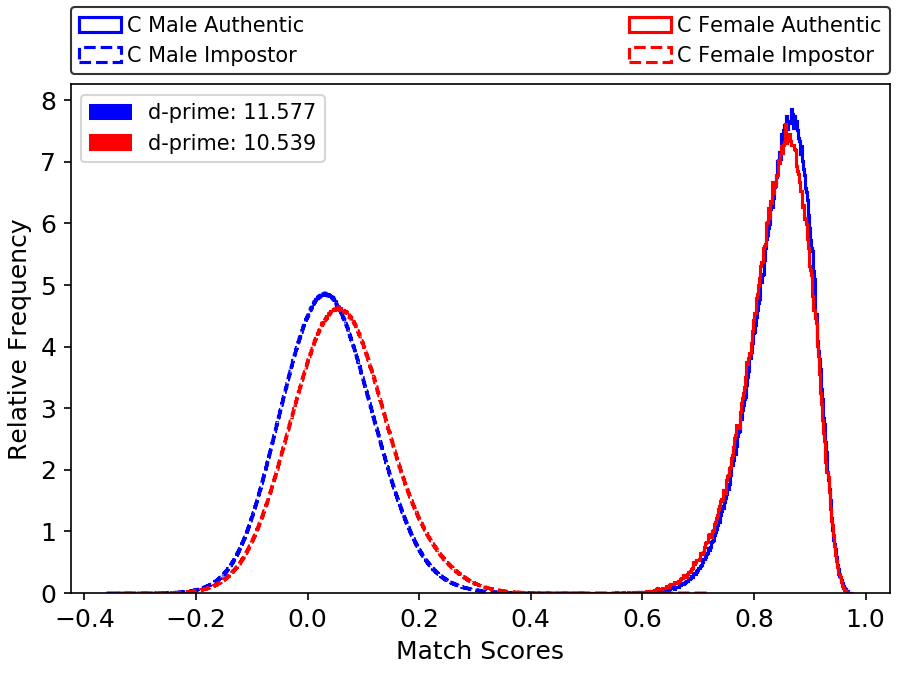}
          \end{subfigure}
          \caption{Notre Dame}
          \vspace{-0.5em}
      \end{subfigure}
      \hfill 
      \begin{subfigure}[b]{0.24\linewidth}
          \begin{subfigure}[b]{1\columnwidth}
            \centering
            \includegraphics[width=\linewidth]{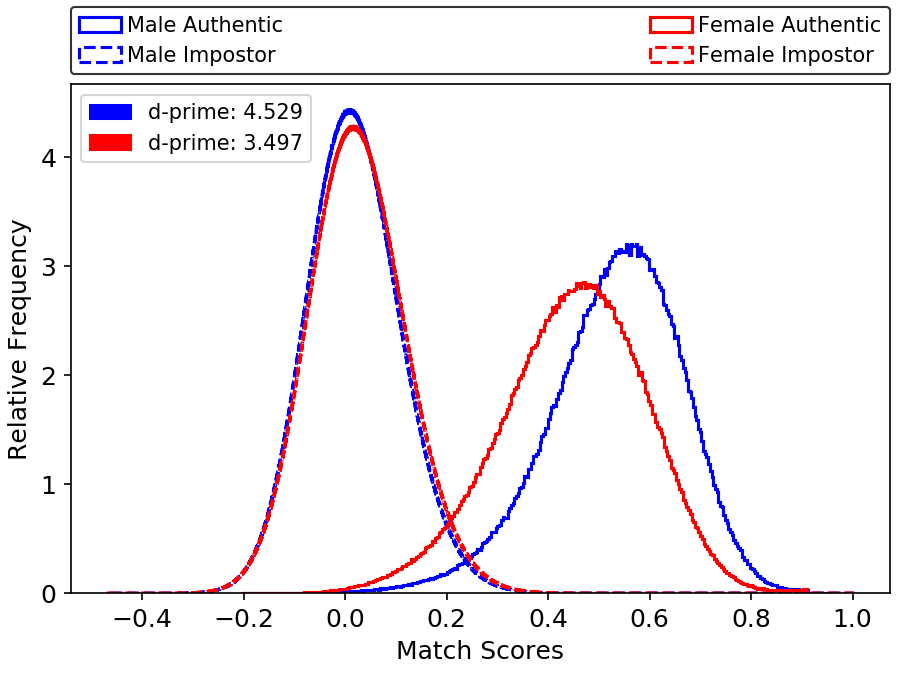}
          \end{subfigure}
          \caption{AFD}
          \vspace{-0.5em}
      \end{subfigure}
  \end{subfigure}
  \caption{Male and female authentic and impostor distributions.}
  \label{fig:aut_imp_original}
  \vspace{-0.5em}
\end{figure*}
\begin{table*}[]
    \setlength\tabcolsep{4pt}
    \centering
    \small
\begin{tabular}{ll|r|r|r|r|r|r}
\multicolumn{1}{l}{} &  & \multicolumn{6}{c}{\textbf{Subset}} \\
\textbf{Dataset} &  & \multicolumn{1}{c|}{\textbf{Original}} & \multicolumn{1}{c|}{\textbf{Neutral Expression}} & \multicolumn{1}{c|}{\textbf{Head Pose A}} & \multicolumn{1}{c|}{\textbf{Head Pose M}} & \multicolumn{1}{c|}{\textbf{Visible Forehead}} & \multicolumn{1}{c}{\textbf{No Makeup}} \\ \hline
\multirow{3}{*}{\textbf{MORPH A-A}} & \textbf{Male} & 10.316 & 10.613 (2.88\%) & 10.45 (1.3\%) & \textbf{10.682 (3.55\%)} & 10.465 (1.44\%) &  \\
 & \textbf{Female} & 8.701 & \textbf{9.083 (4.39\%)} & 8.695 (-0.07\%) & 8.955 (2.92\%) & 8.997 (3.4\%) &  \\
 & \textbf{Diff.} & -1.615 & -1.53 & -1.755 & -1.727 & \textbf{-1.468} &  \\ \hline
\multirow{3}{*}{\textbf{MORPH C}} & \textbf{Male} & 9.641 & 9.908 (2.77\%) & 9.676 (0.36\%) & 9.901 (2.7\%) & \textbf{9.939 (3.09\%)} & 9.641 (Original) \\
 & \textbf{Female} & 8.266 & \textbf{8.874 (7.36\%)} & 8.423 (1.9\%) & 8.612 (4.19\%) & 8.6 (4.04\%) & 8.776 (6.17\%) \\
 & \textbf{Diff.} & -1.375 & -1.034 & -1.253 & -1.289 & -1.339 & \textbf{-0.865} \\ \hline
\multirow{3}{*}{\textbf{Notre Dame}} & \textbf{Male} & 11.577 & \textbf{12.442 (7.47\%)} & 11.561 (-0.14\%) & 11.621 (0.38\%) & 11.93 (3.05\%) & 11.577  (Original)\\
 & \textbf{Female} & 10.539 & \textbf{11.531 (9.41\%)} & 10.722 (1.74\%) & 10.558 (0.18\%) & 10.469 (-0.66\%) & 10.765 (2.14\%) \\
 & \textbf{Diff.} & -1.038 & -0.911 & -0.839 & -1.063 & -1.461 & \textbf{-0.812} \\ \hline
\multirow{3}{*}{\textbf{AFD}} & \textbf{Male} & 4.529 & \textbf{4.81 (6.2\%)} & 4.501 (-0.62\%) & 4.765 (5.21\%) &  & 4.529  (Original) \\
 & \textbf{Female} & 3.497 & \textbf{3.866 (10.55\%)} & 3.455 (-1.2\%) & 3.66 (4.66\%) &  & 3.739 (6.92\%) \\
 & \textbf{Diff.} & -1.032 & -0.944 & -1.046 & -1.105 &  & \textbf{-0.79}
\end{tabular}
    \vspace{-0.5em}
    \caption{Male and female d-prime values, difference (female - male) and percentage improvement (inside parenthesis) from the original dataset. Head poses results are filtered with Amazon (A) and Microsoft (M) face APIs.}
    \vspace{-1em}
    \label{tab:d_prime}
\end{table*}

\subsection{ROC Curve Difference Between Genders}
Figure \ref{fig:roc_original} shows the ROC curves for the four datasets analyzed.
Starting with the three constrained datasets, as all the results are near ceiling, the accuracy of males and females is very close, with the largest gender gap in the MORPH Caucasian dataset.
However, although not usually shown in ROC curves, the thresholds used to achieve specific FMRs for males and females are annotated on the ROCs and are quite different for males and females.
Females have much higher thresholds.
Across the three datasets, the threshold used to achieve a FMR of 1-in-100,000 for males is almost the same as the threshold for a FMR of 1-in-10,000 used for females.
If used in an operational scenario, where usually a fixed threshold is used for all persons, females would have 10 times more false matches than males.
This result suggests two things: (a) when comparing ROC curves between subsets of the same dataset, thresholds should be taken into account; (b) as females have much higher thresholds, their impostor distribution must be shifted to higher scores.

Moving to the unconstrained dataset (AFD), the difference in accuracy between males and females is large, but the thresholds for the same FMR are less different.
As the AFD dataset is collected from the web, the female cohort could have more facial occlusion, makeup, facial expression or larger variation in head pose than the male cohort, which could be responsible for the big difference in accuracy.

\subsection{Impostor and Genuine Distributions}
To better understand the cause of the male/female accuracy difference, Figure \ref{fig:aut_imp_original} shows the genuine and impostor distribution for the four datasets.
Compared to males, the female impostor distribution is shifted towards higher values, and also the female genuine distribution is shifted towards lower values.
For the constrained datasets, the female impostor distribution is shifted by a larger amount than in the unconstrained dataset.
Also, in the unconstrained dataset, the genuine distribution has a larger shift than in the constrained ones.
The d-prime values, which measure the separation between genuine and impostor distribution, are also consistently lower for females.

\begin{figure*}[t]
  \begin{subfigure}[b]{1\linewidth}
      \begin{subfigure}[b]{0.24\linewidth}
          \begin{subfigure}[b]{1\columnwidth}
            \centering
            \includegraphics[width=\linewidth]{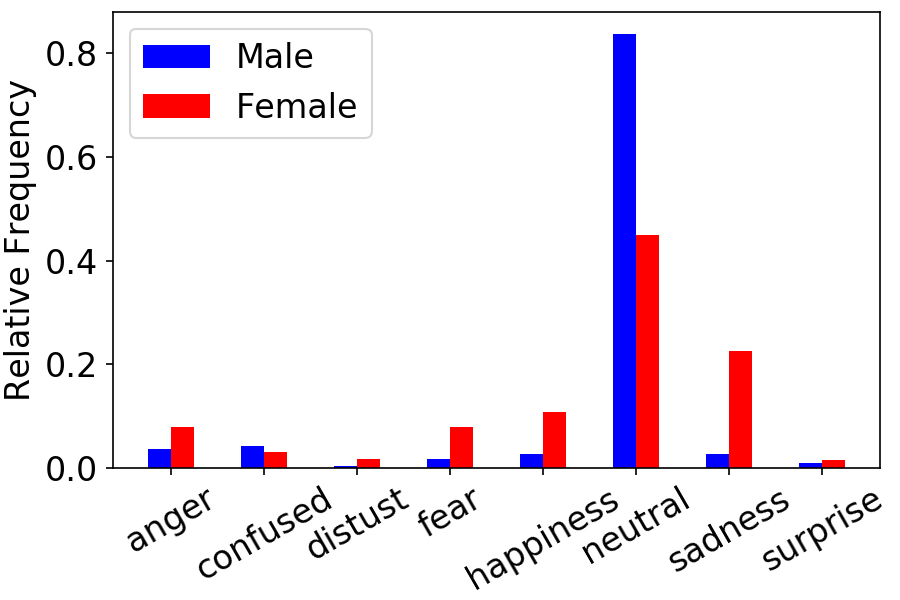}
          \end{subfigure}
      \end{subfigure}
      \hfill 
      \begin{subfigure}[b]{0.24\linewidth}
          \begin{subfigure}[b]{1\columnwidth}
            \centering
            \includegraphics[width=\linewidth]{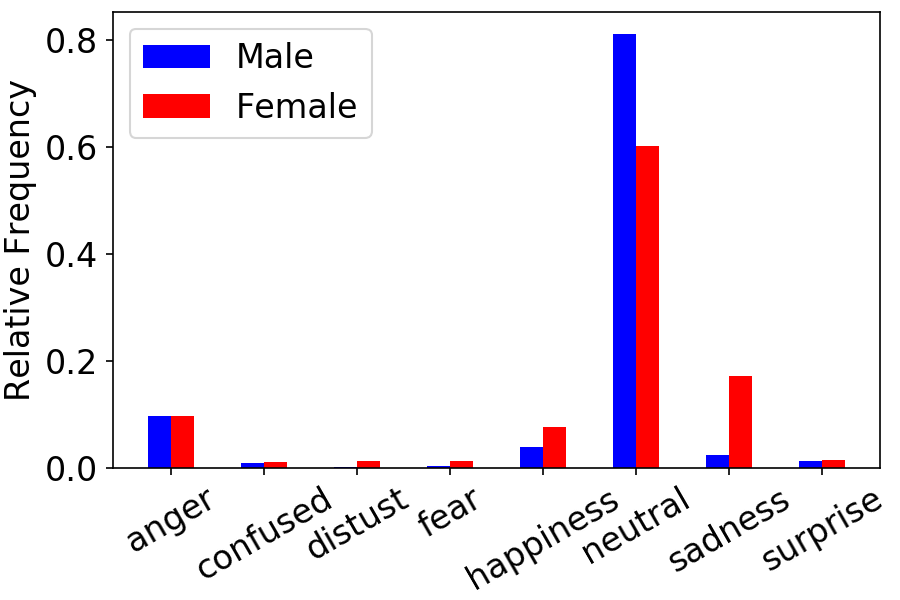}
          \end{subfigure}
      \end{subfigure}
      \hfill 
      \begin{subfigure}[b]{0.24\linewidth}
          \begin{subfigure}[b]{1\columnwidth}
            \centering
            \includegraphics[width=\linewidth]{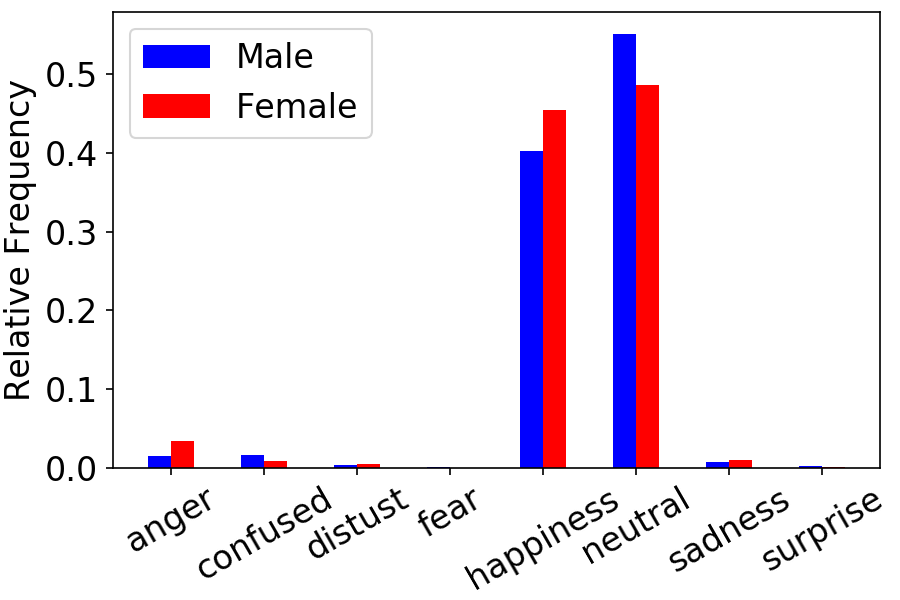}
          \end{subfigure}
      \end{subfigure}
      \hfill 
      \begin{subfigure}[b]{0.24\linewidth}
          \begin{subfigure}[b]{1\columnwidth}
            \centering
            \includegraphics[width=\linewidth]{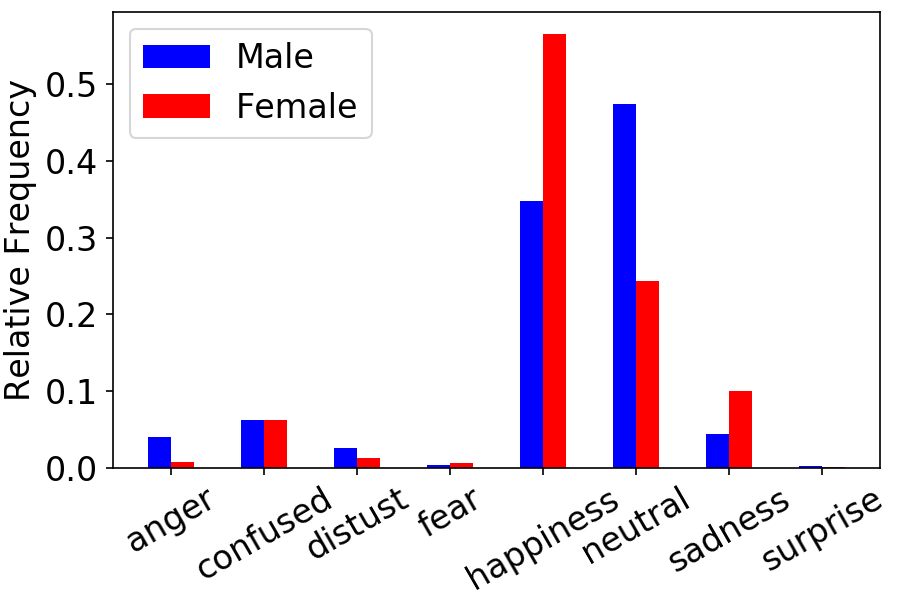}
          \end{subfigure}
      \end{subfigure}
  \end{subfigure}

  \begin{subfigure}[b]{1\linewidth}
      \begin{subfigure}[b]{0.24\linewidth}
          \begin{subfigure}[b]{1\columnwidth}
            \centering
            \includegraphics[width=\linewidth]{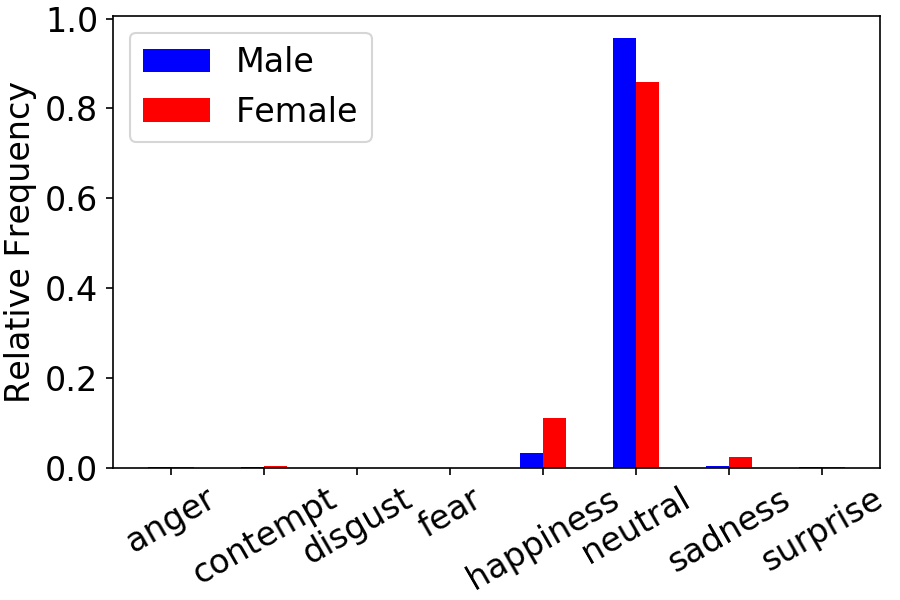}
          \end{subfigure}
          \caption{MORPH African American}
          \vspace{-0.5em}
      \end{subfigure}
      \hfill 
      \begin{subfigure}[b]{0.24\linewidth}
          \begin{subfigure}[b]{1\columnwidth}
            \centering
            \includegraphics[width=\linewidth]{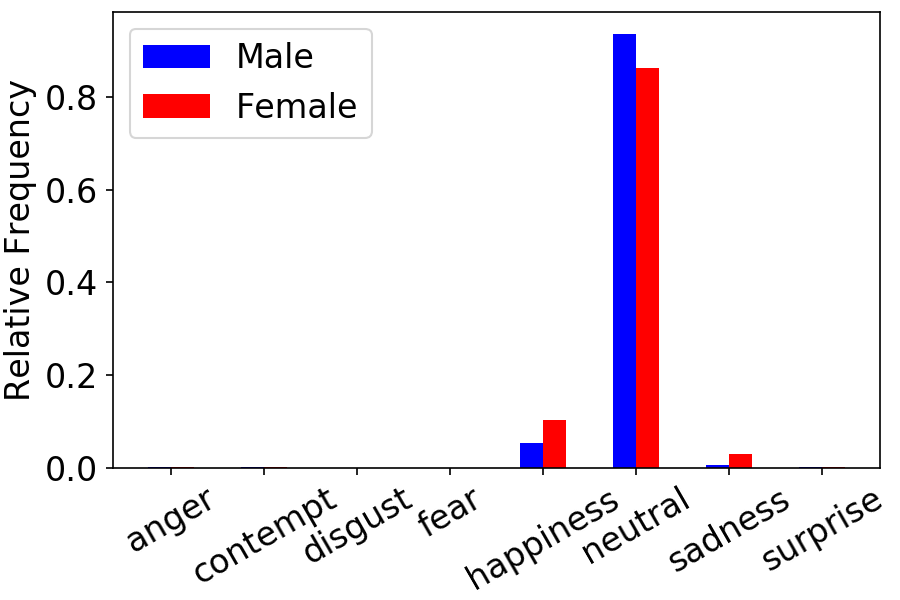}
          \end{subfigure}
          \caption{MORPH Caucasian}
          \vspace{-0.5em}
      \end{subfigure}
      \hfill 
      \begin{subfigure}[b]{0.24\linewidth}
          \begin{subfigure}[b]{1\columnwidth}
            \centering
            \includegraphics[width=\linewidth]{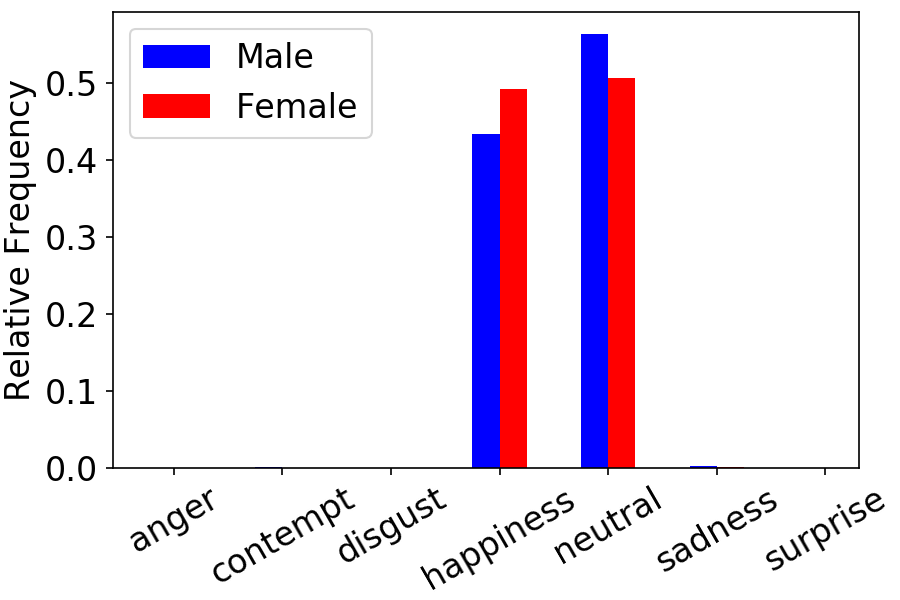}
          \end{subfigure}
          \caption{Notre Dame}
          \vspace{-0.5em}
      \end{subfigure}
      \hfill 
      \begin{subfigure}[b]{0.24\linewidth}
          \begin{subfigure}[b]{1\columnwidth}
            \centering
            \includegraphics[width=\linewidth]{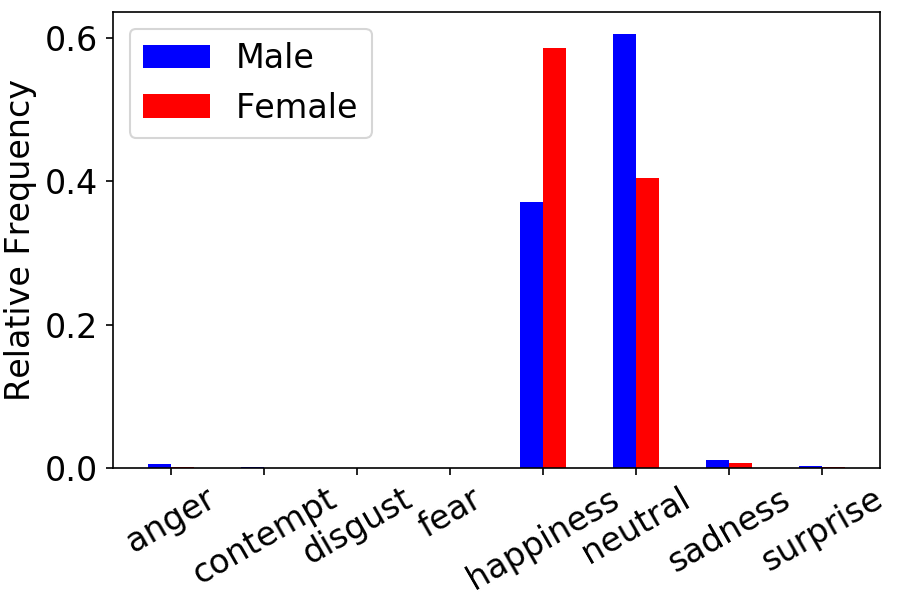}
          \end{subfigure}
          \caption{AFD}
          \vspace{-0.5em}
      \end{subfigure}
  \end{subfigure}
  \caption{Male and female facial expression prediction using Amazon Face API (top) and Microsoft Face API (bottom).}
  \label{fig:emotion_dist}
\end{figure*}
\begin{figure*}[t]
  \begin{subfigure}[b]{1\linewidth}
      \begin{subfigure}[b]{0.24\linewidth}
          \begin{subfigure}[b]{1\columnwidth}
            \centering
            \includegraphics[width=\linewidth]{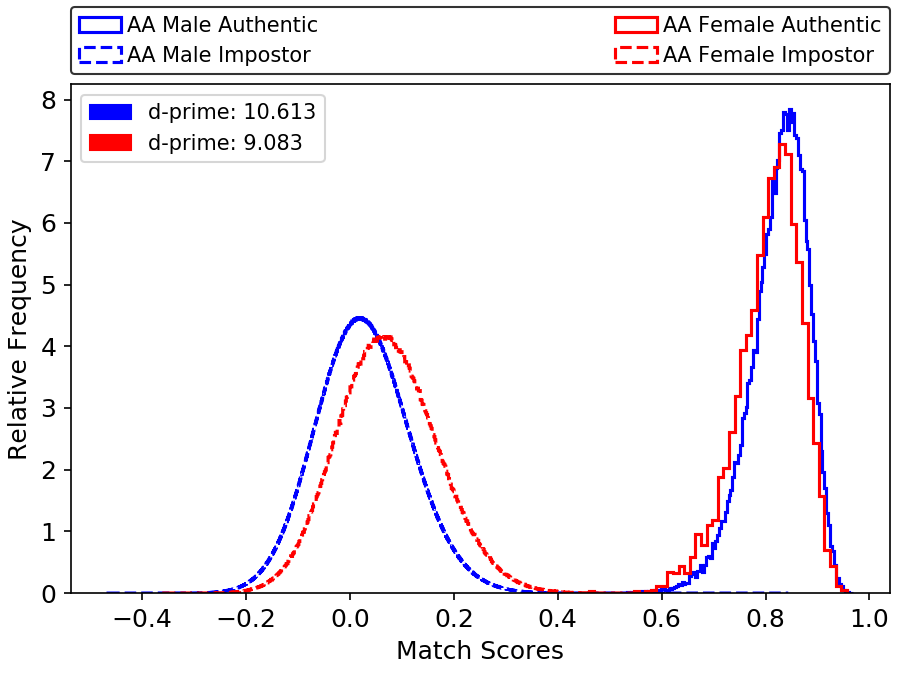}
          \end{subfigure}
          \caption{MORPH African American}
          \vspace{-0.5em}
      \end{subfigure}
      \hfill 
      \begin{subfigure}[b]{0.24\linewidth}
          \begin{subfigure}[b]{1\columnwidth}
            \centering
            \includegraphics[width=\linewidth]{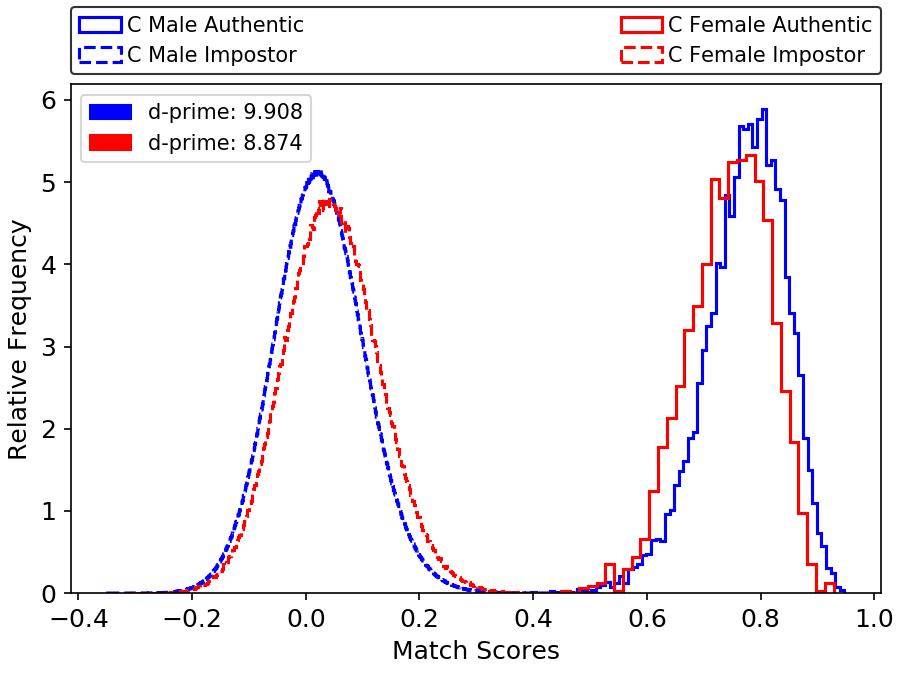}
          \end{subfigure}
          \caption{MORPH Caucasian}
          \vspace{-0.5em}
      \end{subfigure}
      \hfill 
      \begin{subfigure}[b]{0.24\linewidth}
          \begin{subfigure}[b]{1\columnwidth}
            \centering
            \includegraphics[width=\linewidth]{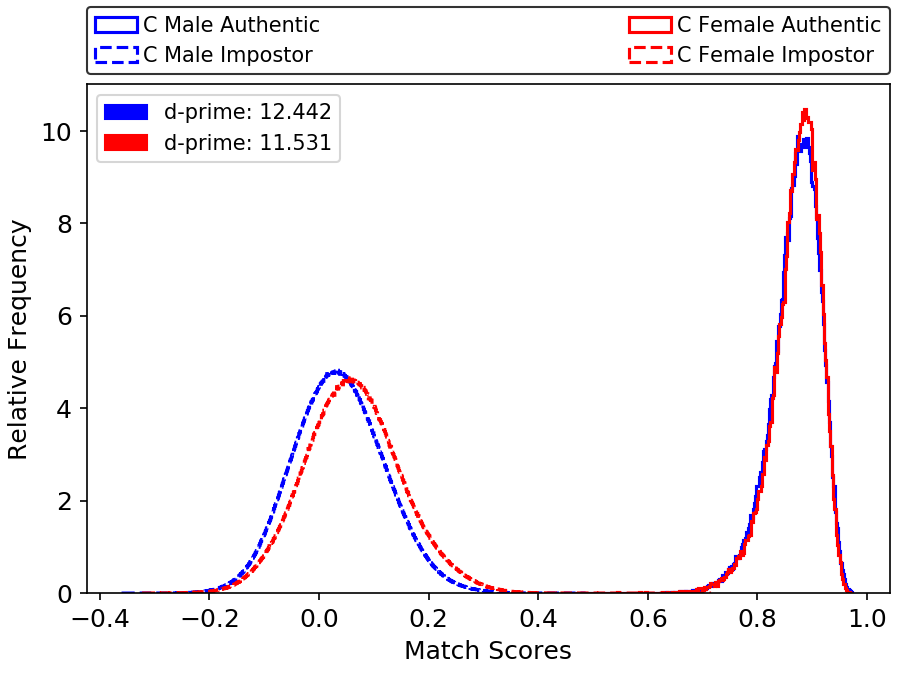}
          \end{subfigure}
          \caption{Notre Dame}
          \vspace{-0.5em}
      \end{subfigure}
      \hfill 
      \begin{subfigure}[b]{0.24\linewidth}
          \begin{subfigure}[b]{1\columnwidth}
            \centering
            \includegraphics[width=\linewidth]{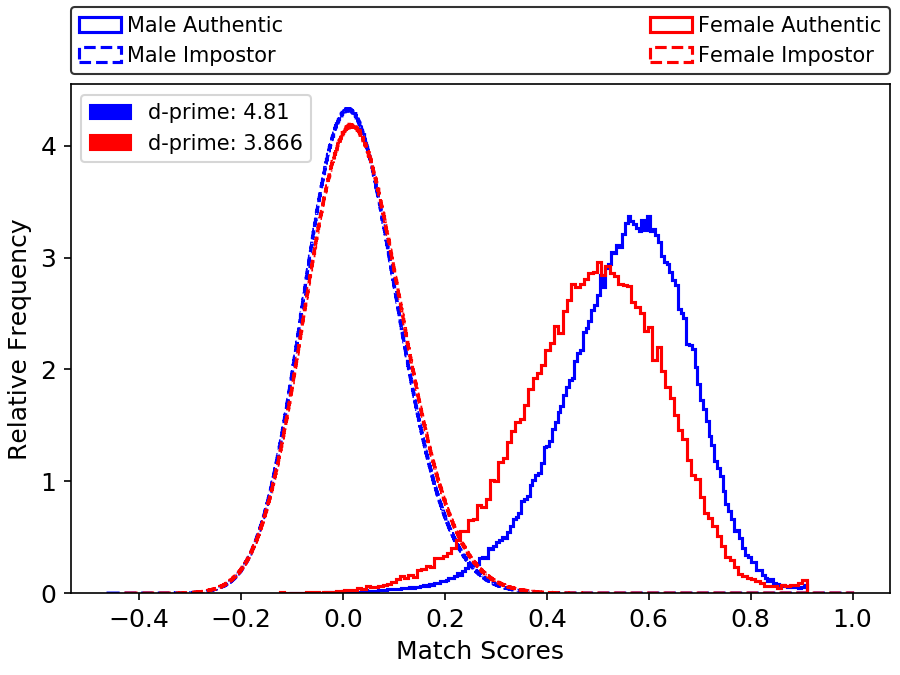}
          \end{subfigure}
          \caption{AFD}
          \vspace{-0.5em}
      \end{subfigure}
  \end{subfigure}
  \caption{Male and female authentic and impostor distributions using only images with neutral facial expression.}
  \label{fig:aut_imp_emotion}
  \vspace{-1em}
\end{figure*}

\subsection{Investigating the Causes}
In this section, we test some of the speculated causes for female face recognition accuracy to be lower than for males.
The d-prime values achieved for males and females after each speculated cause is controlled for are compared to the original values in Table \ref{tab:d_prime}.

\subsubsection{Facial Expression}
One of the speculated causes of lower accuracy for females is that females exhibit a broader range of facial expression when photographed, which could lead to lower similarity scores for genuine image pairs.
To validate if women indeed have more non-neutral images than men, we used two commercial facial analysis APIs to predict facial expression for the images.

Starting with the Notre Dame dataset, which has meta-data for the expression prompted at image acquisition, Amazon Face API agreed with 91.79\% (6911/7529) of the male neutral expression labels, and  86.32\% (4480/5190) of the female.
The Microsoft Face API agreed with 96.67\% (7278/7529) of the male neutral expression labels, and 93.93\% (4875/5190) of the female.
We analyze the images for which either API did not agree with the mta-data, and indeed many were not neutral expression.
Thus, to create a highly-confident neutral expression subset, we selected images that all three sources (meta-data labels, Amazon Face API, and Microsoft Face API) classified as neutral expression, giving us 6,858 male images, and 4,361 females images, for the same set of subjects.

\begin{figure*}[t]

  \begin{subfigure}[b]{1\linewidth}
      \begin{subfigure}[b]{0.24\linewidth}
          \begin{subfigure}[b]{1\columnwidth}
            \centering
            \includegraphics[width=\linewidth]{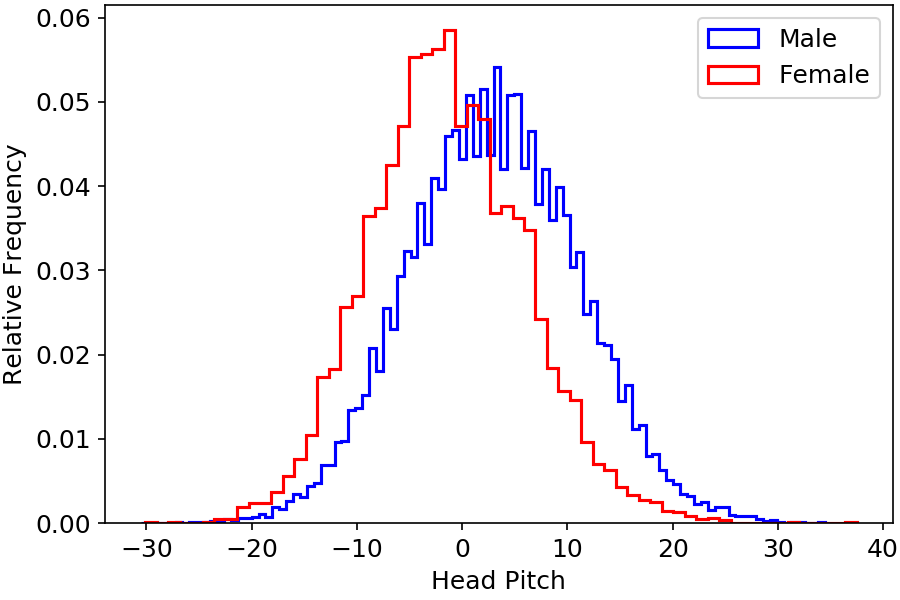}
          \end{subfigure}
          \caption{MORPH African American}
          \vspace{-0.5em}
      \end{subfigure}
      \hfill 
      \begin{subfigure}[b]{0.24\linewidth}
          \begin{subfigure}[b]{1\columnwidth}
            \centering
            \includegraphics[width=\linewidth]{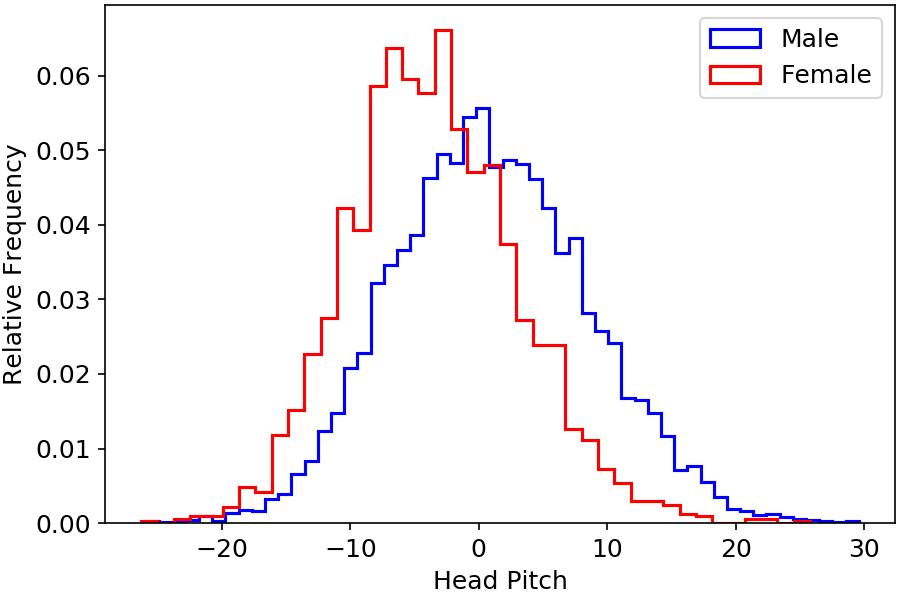}
          \end{subfigure}
          \caption{MORPH Caucasian}
          \vspace{-0.5em}
      \end{subfigure}
      \hfill 
      \begin{subfigure}[b]{0.24\linewidth}
          \begin{subfigure}[b]{1\columnwidth}
            \centering
            \includegraphics[width=\linewidth]{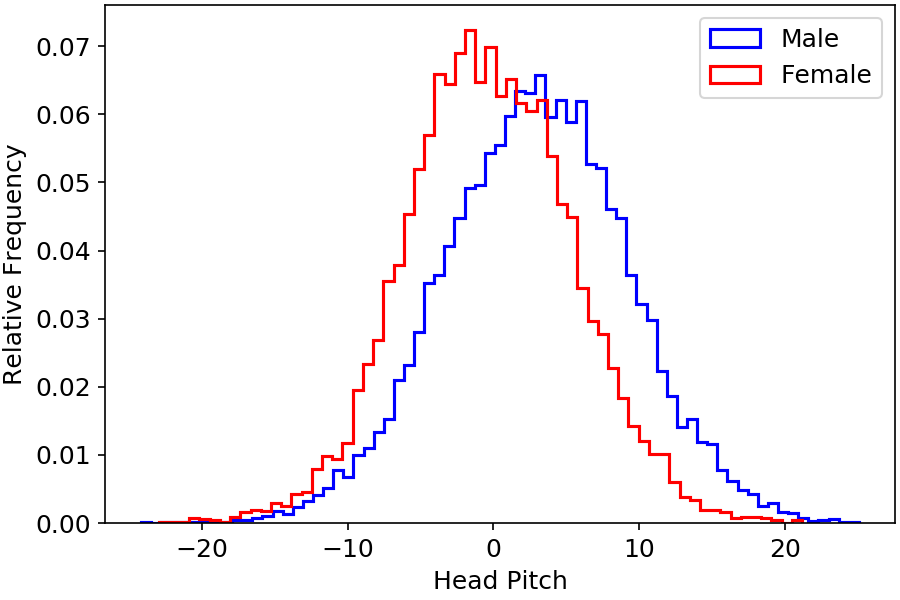}
          \end{subfigure}
          \caption{Notre Dame}
          \vspace{-0.5em}
      \end{subfigure}
      \hfill 
      \begin{subfigure}[b]{0.24\linewidth}
          \begin{subfigure}[b]{1\columnwidth}
            \centering
            \includegraphics[width=\linewidth]{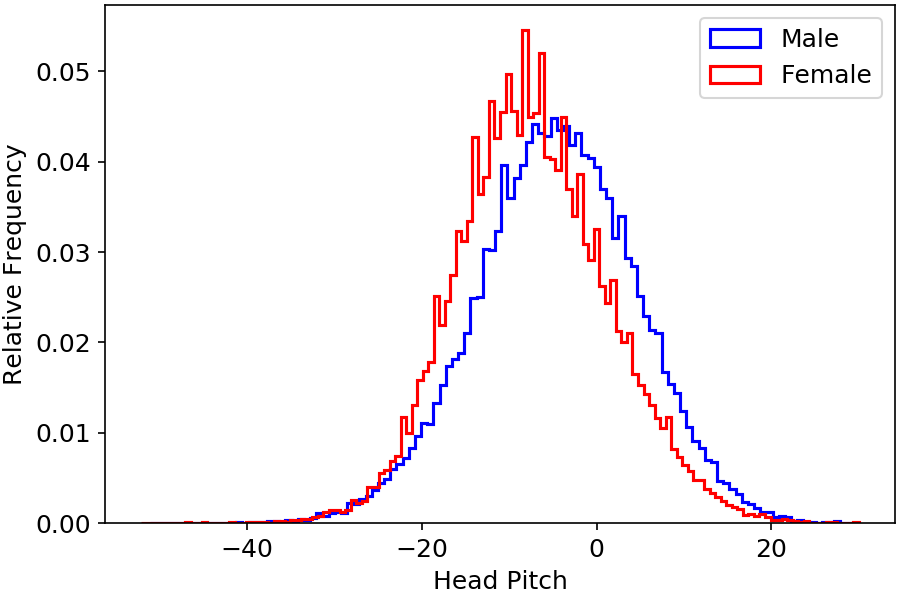}
          \end{subfigure}
          \caption{AFD}
          \vspace{-0.5em}
      \end{subfigure}
  \end{subfigure}
  \caption{Male and female head pose pitch predicted using Microsoft Face API.}
  \label{fig:pitch_dist}
\end{figure*}
\begin{figure*}[t]
  \begin{subfigure}[b]{1\linewidth}
      \begin{subfigure}[b]{0.24\linewidth}
          \begin{subfigure}[b]{1\columnwidth}
            \centering
            \includegraphics[width=\linewidth]{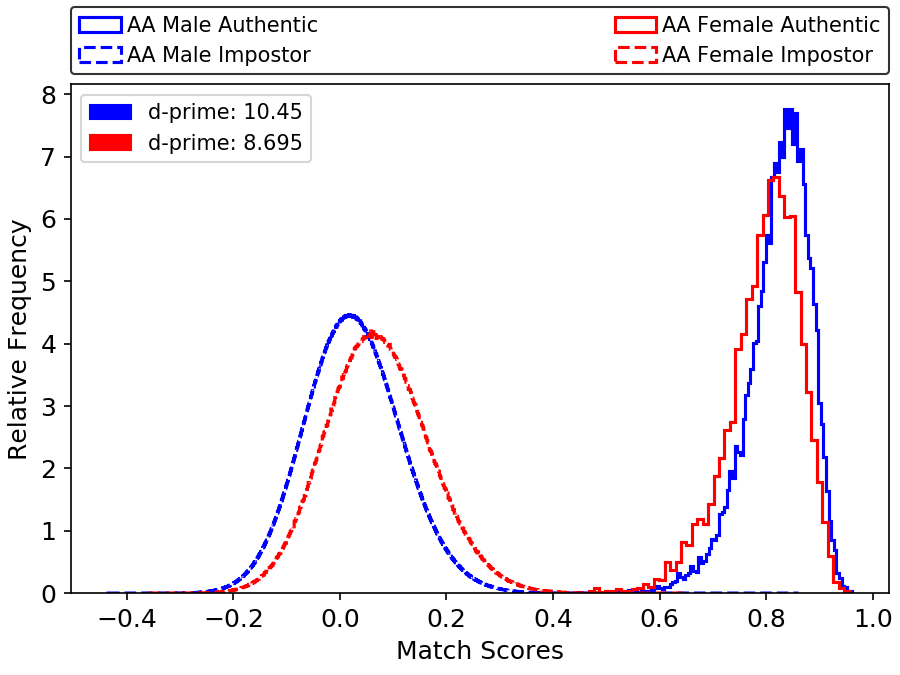}
          \end{subfigure}
      \end{subfigure}
      \hfill 
      \begin{subfigure}[b]{0.24\linewidth}
          \begin{subfigure}[b]{1\columnwidth}
            \centering
            \includegraphics[width=\linewidth]{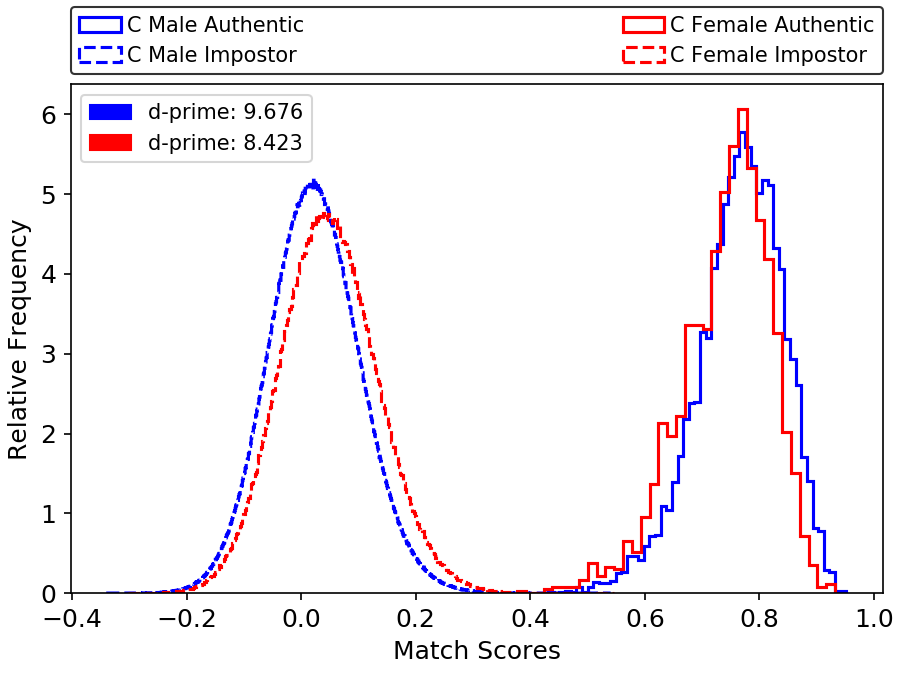}
          \end{subfigure}
      \end{subfigure}
      \hfill 
      \begin{subfigure}[b]{0.24\linewidth}
          \begin{subfigure}[b]{1\columnwidth}
            \centering
            \includegraphics[width=\linewidth]{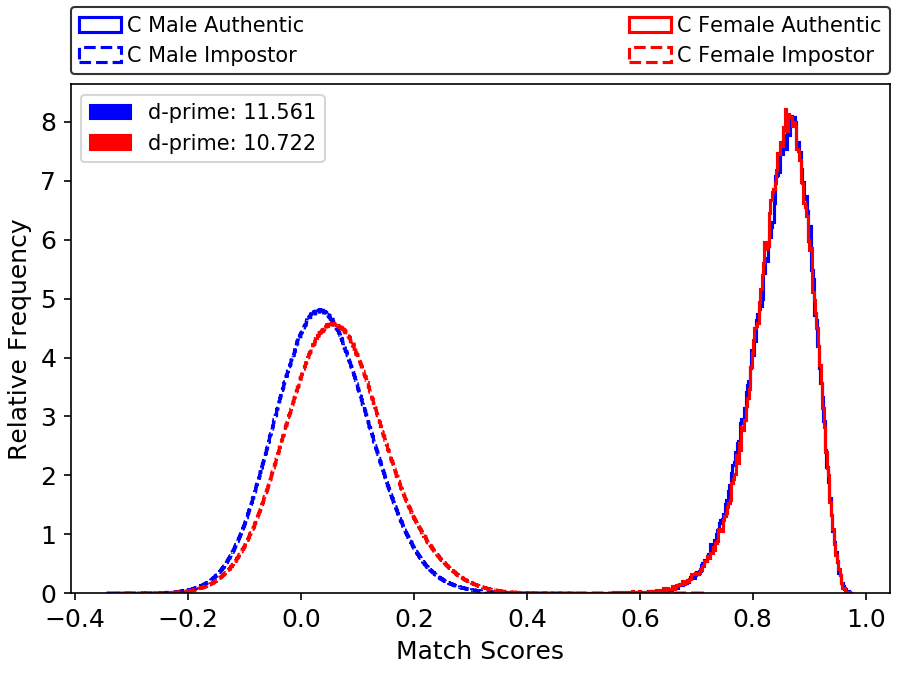}
          \end{subfigure}
      \end{subfigure}
      \hfill 
      \begin{subfigure}[b]{0.24\linewidth}
          \begin{subfigure}[b]{1\columnwidth}
            \centering
            \includegraphics[width=\linewidth]{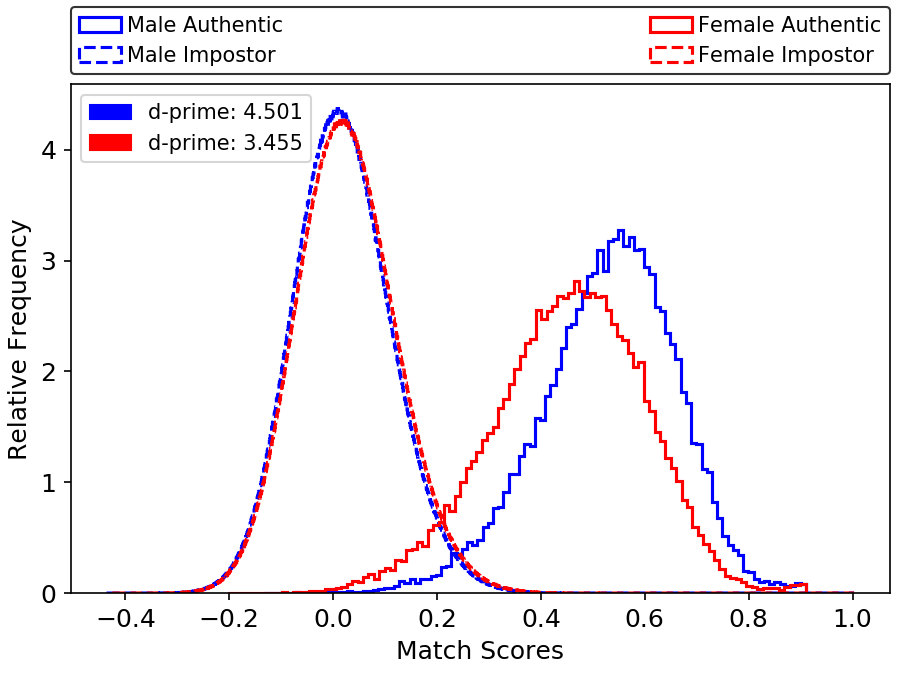}
          \end{subfigure}
      \end{subfigure}
  \end{subfigure}
  
  \begin{subfigure}[b]{1\linewidth}
      \begin{subfigure}[b]{0.24\linewidth}
          \begin{subfigure}[b]{1\columnwidth}
            \centering
            \includegraphics[width=\linewidth]{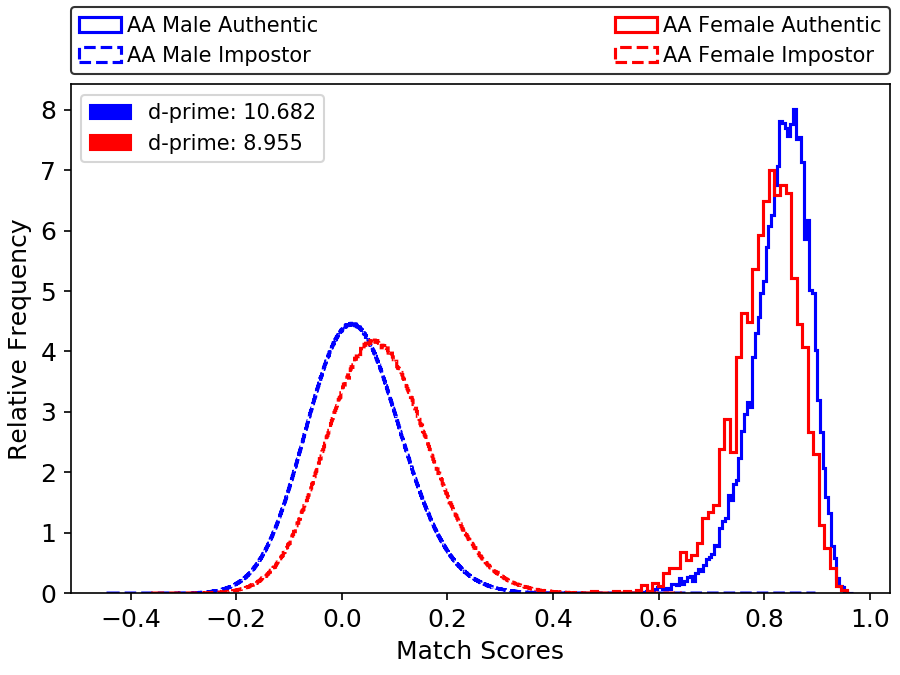}
          \end{subfigure}
          \caption{MORPH African American}
          \vspace{-0.5em}
      \end{subfigure}
      \hfill 
      \begin{subfigure}[b]{0.24\linewidth}
          \begin{subfigure}[b]{1\columnwidth}
            \centering
            \includegraphics[width=\linewidth]{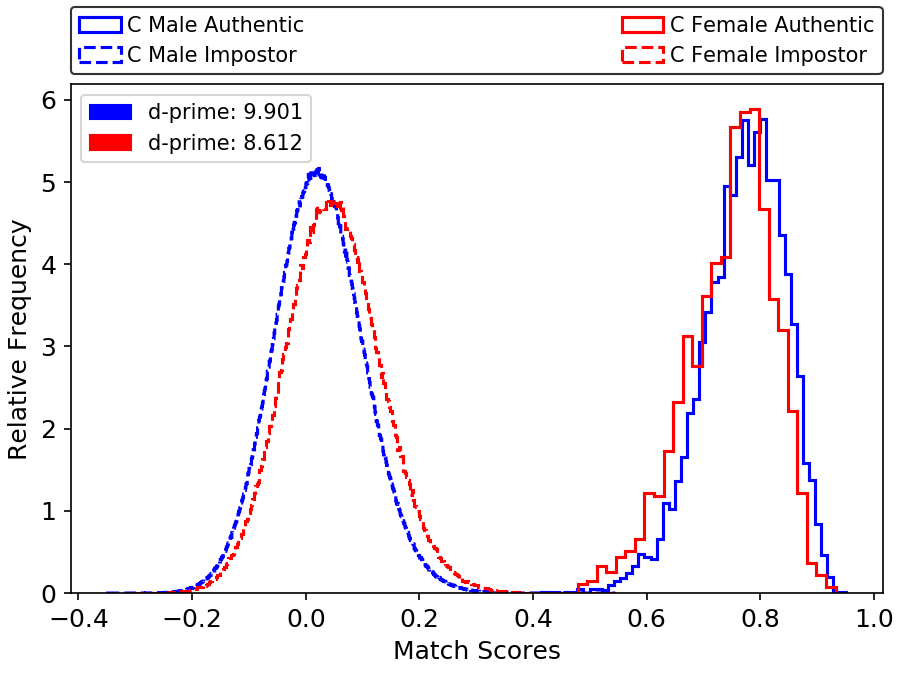}
          \end{subfigure}
          \caption{MORPH Caucasian}
          \vspace{-0.5em}
      \end{subfigure}
      \hfill 
      \begin{subfigure}[b]{0.24\linewidth}
          \begin{subfigure}[b]{1\columnwidth}
            \centering
            \includegraphics[width=\linewidth]{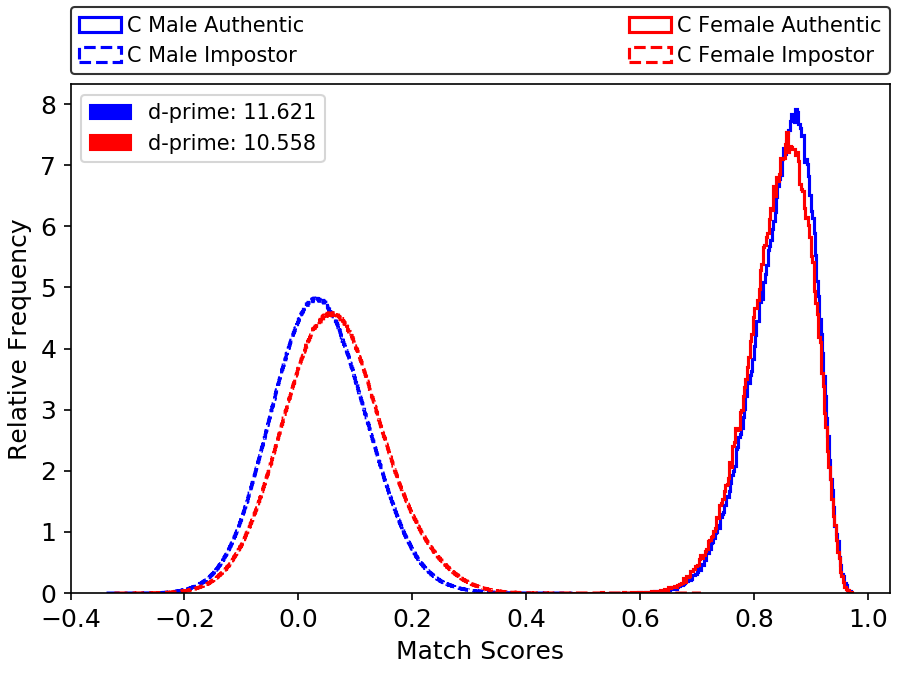}
          \end{subfigure}
          \caption{Notre Dame}
          \vspace{-0.5em}
      \end{subfigure}
      \hfill 
      \begin{subfigure}[b]{0.24\linewidth}
          \begin{subfigure}[b]{1\columnwidth}
            \centering
            \includegraphics[width=\linewidth]{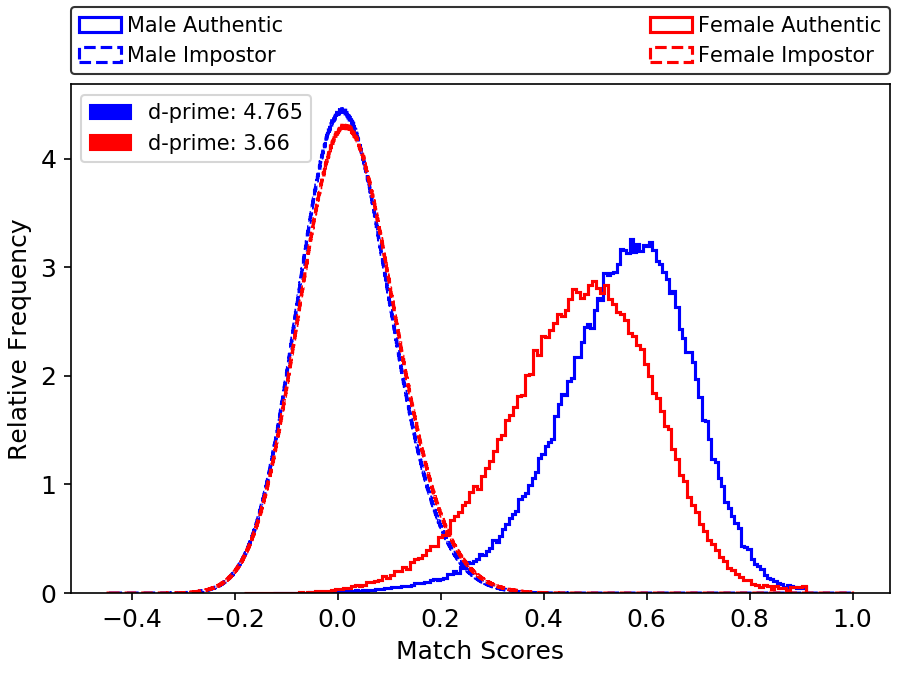}
          \end{subfigure}
          \caption{AFD}
          \vspace{-0.5em}
      \end{subfigure}
  \end{subfigure}
  \caption{Male and female authentic and impostor distributions using only images with -5 and 5 degrees of head pitch angle, filtered using Amazon Face API (top) and Microsoft Face API (bottom).}
  \label{fig:aut_imp_pitch}
  \vspace{-1em}
\end{figure*}

MORPH and AFD do not have meta-data for expression, so we use images rated as neutral by both APIs. 
For MORPH African-American, the male cohort had 30,395 images rated as neutral by both APIs, and the female cohort had 2,529.
On MORPH Caucasian, 6,362 male images and 1,491 females images were rated as neutral by both APIs.
Finally, on the AFD dataset, 17,469 male and 10,900 female images were classified as neutral facial expression by both APIs.

Figure \ref{fig:emotion_dist} shows the distribution of facial expression using both face APIs.
Males show consistently more neutral facial expression than females across all four datasets.
In the same way, females have more images with a ``happy'' expression, which agrees with the speculation that females smile more at time of image acquisition.

Figure \ref{fig:aut_imp_emotion} shows the genuine and impostor distribution for the datasets using only images classified with neutral facial expression.
As shown in Table \ref{tab:d_prime}, the separation between the impostor and genuine distribution increased for both males and females, but females had a larger increase than males, which was expected since females had more non-neutral facial expression images removed.
The highest increase in d-prime for males is 7.47\% in the Notre Dame dataset, and for females is 10.55\% in the AFD dataset.
However, except for the Notre Dame dataset, where the female genuine distribution is slightly shifted towards higher (better) values than the male genuine distribution, the previously-seen pattern holds, and both genuine and impostor distributions are still shifted towards each other compared to males.

\subsubsection{Head Pose}

Another speculated cause of lower accuracy for women is that they might have a greater incidence of off-angle pose in the images.
This could occur if a camera is adjusted to an average male height, and then used to acquire images of women and men (as speculated in \cite{cook2018}).

Using the two commercial face APIs, we detected the head pose pitch of males and females, which is shown in Figure \ref{fig:pitch_dist} for one of them, due to space constrains.
For all datasets, the male and female pitch distributions are similar. However, off-angle pose could potentially affect more females than males.

We filtered the images to ones that are within -5 to +5 degrees of head pose pitch.
As there was low agreement between the Amazon and Microsoft Face APIs, we did not combine both into a single subset.
The Amazon Face API classified in the range of -5 and 5 degrees of pitch: 19,080 male images and 3,336 female images on MORPH African-American, 4,585 male images and 1,519 female images on MORPH Caucasian, 6,769 male images and 5,426 female images on Notre Dame, and 7,186 male images and 8,405 female images on AFD.
The Microsoft Face API classified in the range of -5 and 5 degrees of pitch: 16,097 male images and 2,847 female images on MORPH African-American, 3,847 male images and 1,227 female images on MORPH Caucasian, 7,487 male images and 6,283 female images on Notre Dame, and 15,217 male images and 14,229 female images on AFD.
From manually examining some images, the Microsoft Face API seems, for the versions of the APIs and the datasets used here,
to be more accurate on head pose pitch than the Amazon API.

Figure \ref{fig:aut_imp_pitch} shows the male and female genuine and impostor distribution for the images filtered to be within -5 to +5 degrees of head pose pitch.
For the images filtered with Amazon Face API, overall, both males and females had slightly worse genuine and impostor distributions (see Table \ref{tab:d_prime}).
As for Microsoft Face API filtered images, the separation between both male and female distributions had only a small improvement (see Table \ref{tab:d_prime}).
It is expected that images with less variation in head pose will have better matching,
thus this result agrees with our analysis that Microsoft is more accurate to predict head pose pitch angles.
Finally, we still see the same pattern of accuracy difference between males and females after restricting the dataset on pose.

\subsubsection{Forehead Occlusion}
\begin{figure}[t]
  \begin{subfigure}[b]{1\linewidth}
      \begin{subfigure}[b]{0.47\linewidth}
          \begin{subfigure}[b]{.49\columnwidth}
            \centering
            \includegraphics[width=\linewidth]{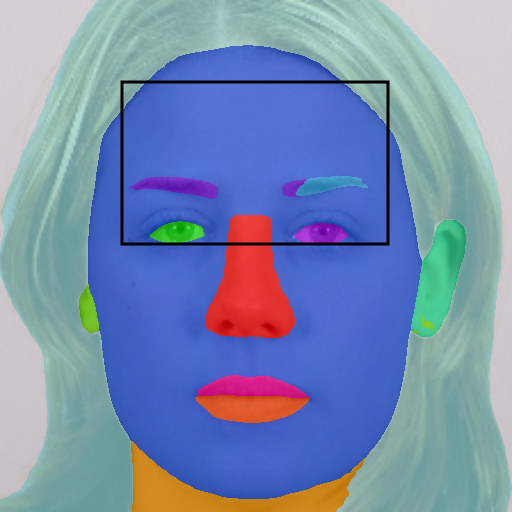}
          \end{subfigure}
          \hfill 
          \begin{subfigure}[b]{.49\columnwidth}
            \centering
            \includegraphics[width=\linewidth]{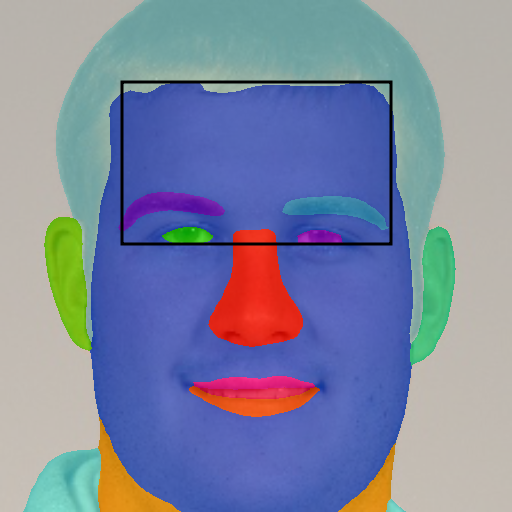}
          \end{subfigure}
      \end{subfigure}
      \hfill 
      \begin{subfigure}[b]{0.47\linewidth}
          \begin{subfigure}[b]{.49\columnwidth}
            \centering
            \includegraphics[width=\linewidth]{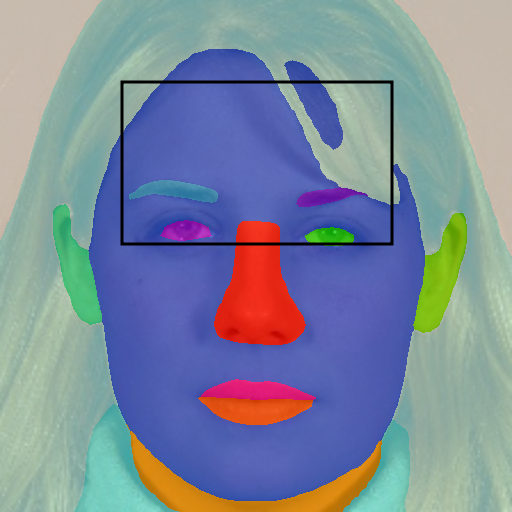}
          \end{subfigure}
          \hfill 
          \begin{subfigure}[b]{.49\columnwidth}
            \centering
            \includegraphics[width=\linewidth]{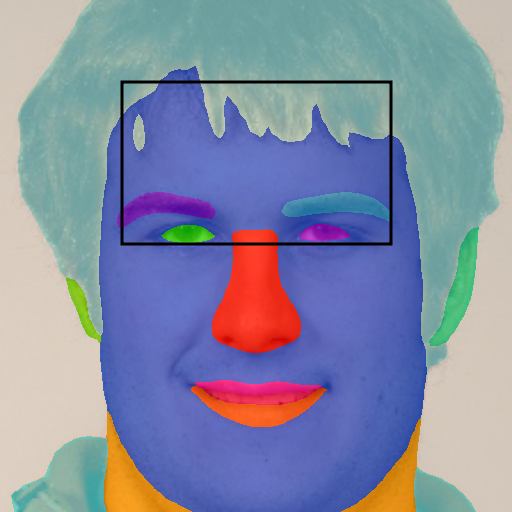}
          \end{subfigure}
      \end{subfigure}
  \end{subfigure}
  \caption{Examples of segmented images that passed (left) and did not passed (right) the forehead occlusion check. Subjects shown in the left and right are the same persons.}
  \label{fig:forehead_examples}
  \vspace{-1em}
\end{figure}

Another speculated cause for female accuracy being lower is that females more frequently have hair occluding part of their face, in particular the forehead and eyes.
To detect if the subjects have their forehead occluded, we used the BiSeNet \cite{bisenet} image segmentation network, that was trained using the CelebAMask-HQ \cite{maskgan}, to segment the face images.
Then, we cropped a region corresponding to the forehead and checked the ratio of hair/hat to other face parts (skin, eyebrows, eyes, and nose) in this region.
All the images that had less than 10\% hair/hat were kept in a forehead-occlusion-free subset.
We filtered both MORPH and Notre Dame datasets, which resulted in: 6,105 images of 1,899 males and 1,501 images of 476 females on MORPH Caucasian; 32,120 images of 8,577 males and 4,574 images of 1,404 females on MORPH African-American; and 9,144 images of 216 males and 6,713 images of 153 females on Notre Dame.
We were not able to repeat the same filtering for AFD dataset, as the faces are not as frontal.
Figure \ref{fig:forehead_examples} shows examples of segmented image faces that passed and did not pass the skin check.

\begin{figure*}[t]
  \begin{subfigure}[b]{1\linewidth}
      \begin{subfigure}[b]{0.27\linewidth}
          \begin{subfigure}[b]{1\columnwidth}
            \centering
            \includegraphics[width=\linewidth]{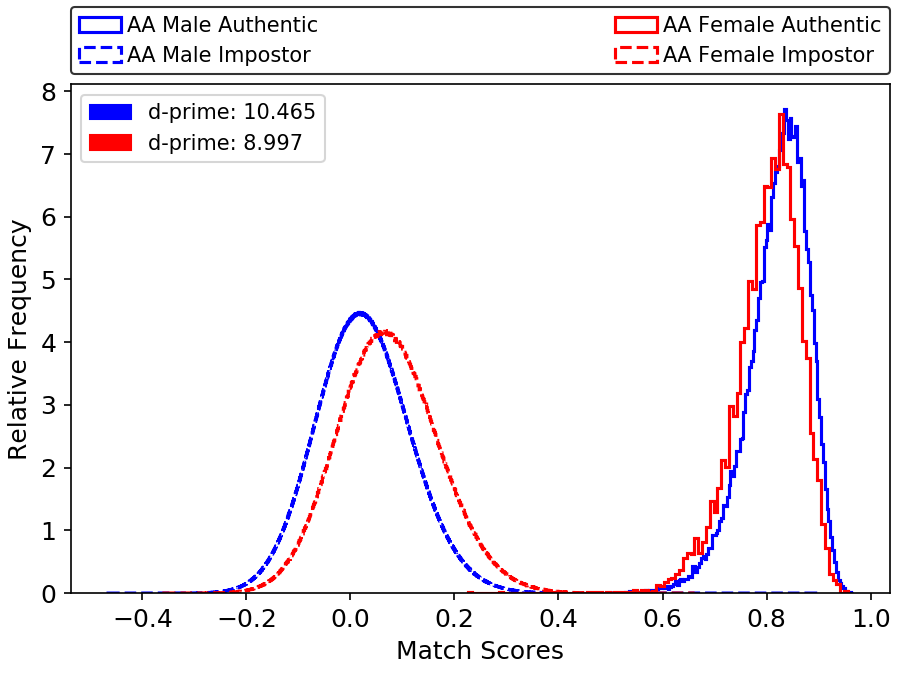}
          \end{subfigure}
          \caption{MORPH African American}
          \vspace{-0.5em}
      \end{subfigure}
      \hfill 
      \begin{subfigure}[b]{0.27\linewidth}
          \begin{subfigure}[b]{1\columnwidth}
            \centering
            \includegraphics[width=\linewidth]{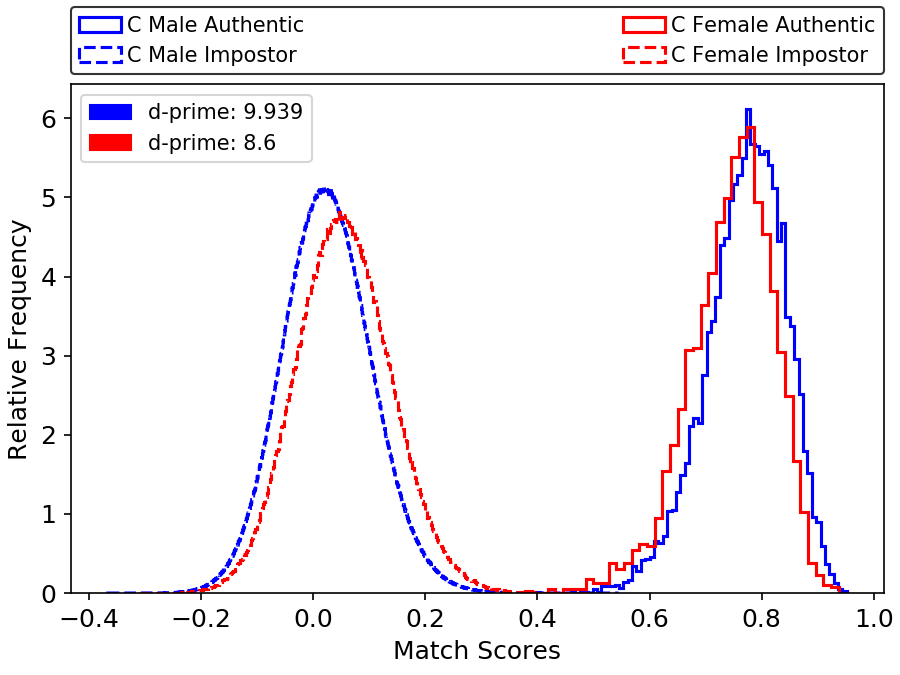}
          \end{subfigure}
          \caption{MORPH Caucasian}
          \vspace{-0.5em}
      \end{subfigure}
      \hfill 
      \begin{subfigure}[b]{0.27\linewidth}
          \begin{subfigure}[b]{1\columnwidth}
            \centering
            \includegraphics[width=\linewidth]{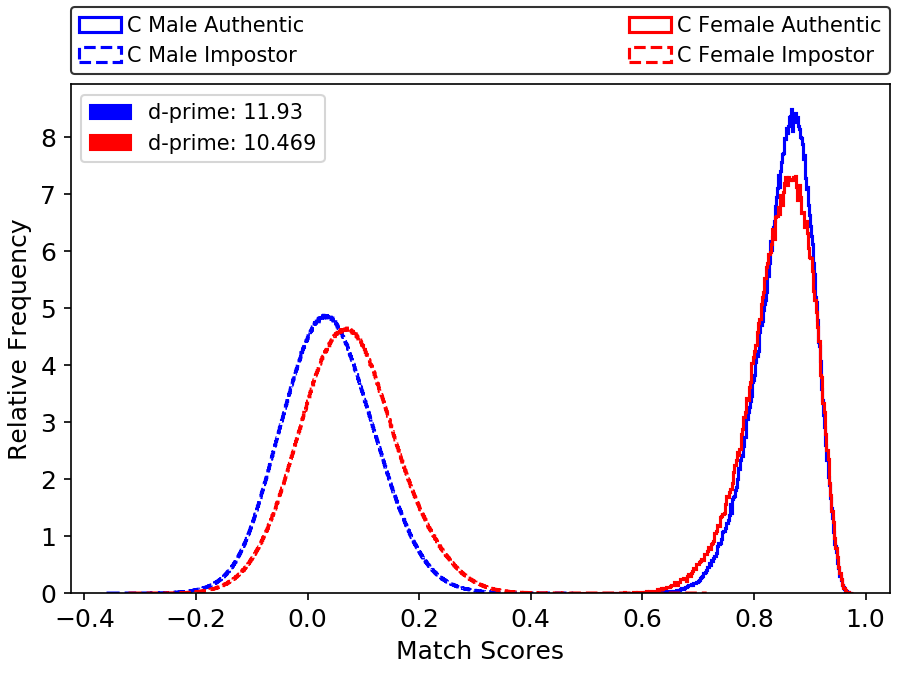}
          \end{subfigure}
          \caption{Notre Dame}
          \vspace{-0.5em}
      \end{subfigure}
  \end{subfigure}
  \caption{Male and female authentic and impostor distributions using only images with no forehead occlusion.}
  \label{fig:aut_imp_forehead}
\end{figure*}
\begin{figure*}[t]
  \begin{subfigure}[b]{1\linewidth}
      \begin{subfigure}[b]{0.27\linewidth}
          \begin{subfigure}[b]{1\columnwidth}
            \centering
            \includegraphics[width=\linewidth]{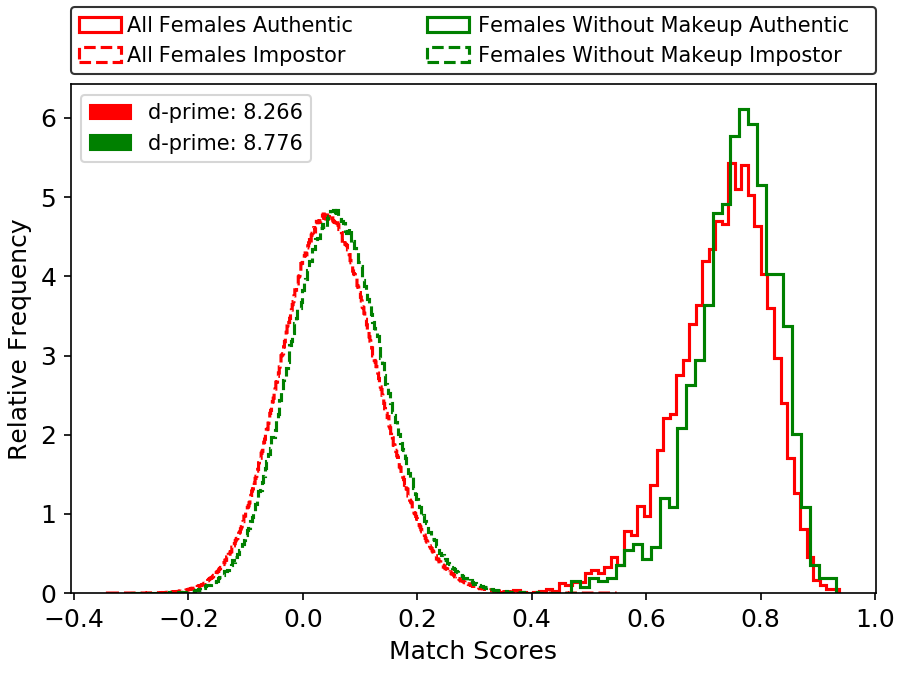}
          \end{subfigure}
          \caption{MORPH Caucasian}
          \vspace{-0.5em}
      \end{subfigure}
      \hfill 
      \begin{subfigure}[b]{0.27\linewidth}
          \begin{subfigure}[b]{1\columnwidth}
            \centering
            \includegraphics[width=\linewidth]{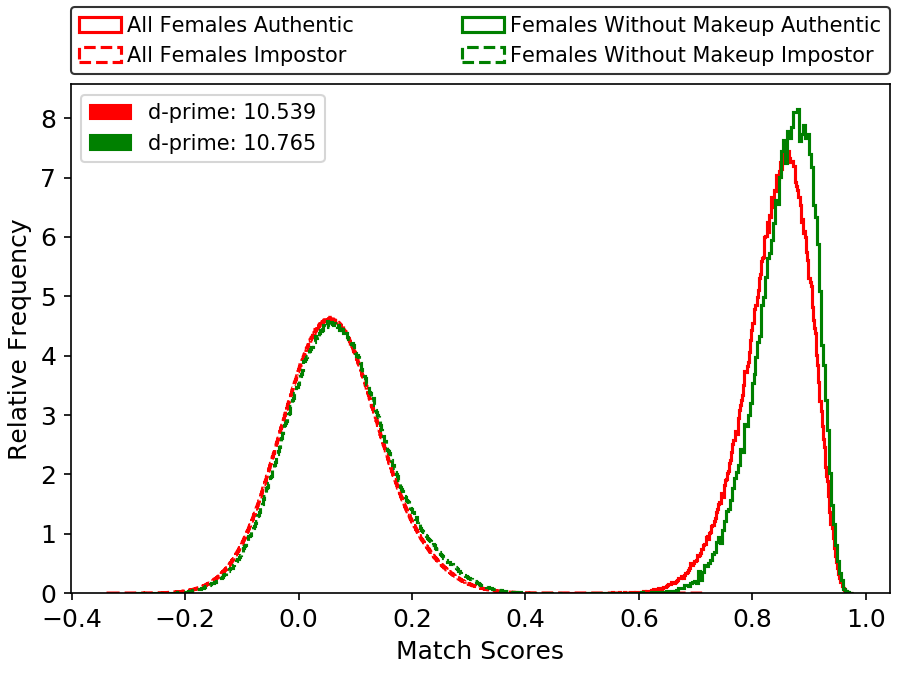}
          \end{subfigure}
          \caption{Notre Dame}
          \vspace{-0.5em}
      \end{subfigure}
      \hfill 
      \begin{subfigure}[b]{0.27\linewidth}
          \begin{subfigure}[b]{1\columnwidth}
            \centering
            \includegraphics[width=\linewidth]{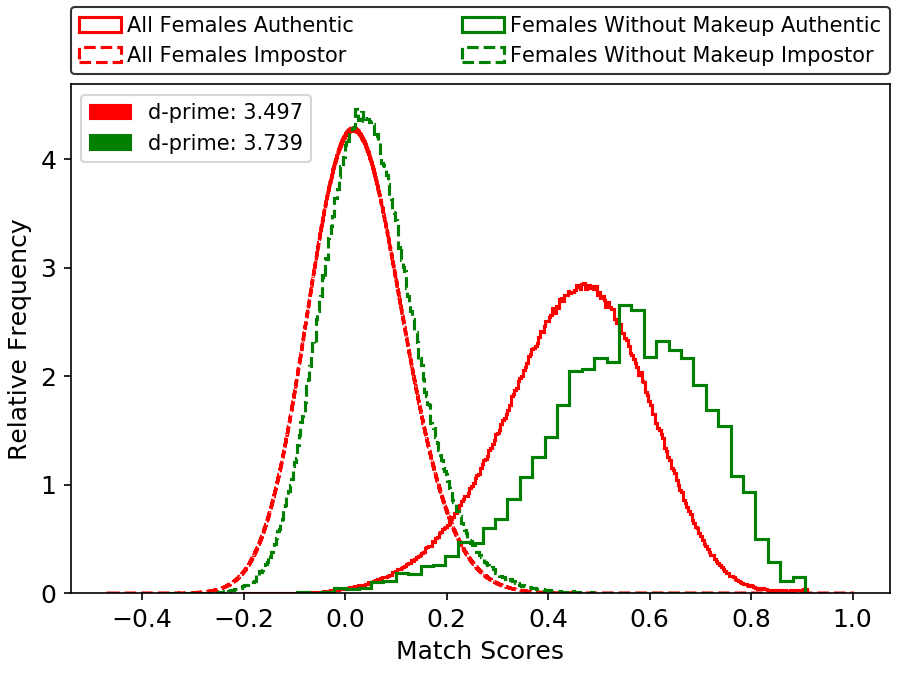}
          \end{subfigure}
          \caption{AFD}
          \vspace{-0.5em}
      \end{subfigure}
  \end{subfigure}
  \caption{Female authentic and impostor distributions with and without makeup.}
  \label{fig:aut_imp_makeup}
  \vspace{-1em}
\end{figure*}

Figure \ref{fig:aut_imp_forehead} shows the male and female genuine and impostor distribution using only images with no occlusion of the forehead.
As shown in Table \ref{tab:d_prime}, except for Notre Dame females, removing forehead occlusion improves the d-prime of both genders.
The highest improvement of both male and female d-prime is MOPRH Caucasian, with 3\% and 3.88\% improvement, respectively.
Although this improved the separation of genuine and impostor distributions for both genders, 
removing forehead occlusion did not make the impostor or genuine distributions more similar across gender. 

\subsubsection{Facial Makeup}
\begin{figure}[t]
  \begin{subfigure}[b]{1\linewidth}
      \begin{subfigure}[b]{1\linewidth}
          \begin{subfigure}[b]{0.25\columnwidth}
            \centering
            \includegraphics[width=\linewidth]{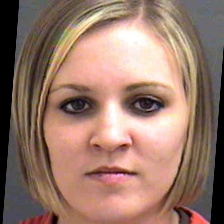}          \end{subfigure}
          \hfill 
          \begin{subfigure}[b]{0.25\columnwidth}
            \centering
            \includegraphics[width=\linewidth]{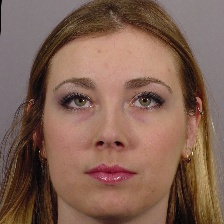}
          \end{subfigure}
          \hfill 
          \begin{subfigure}[b]{0.25\columnwidth}
            \centering
            \includegraphics[width=\linewidth]{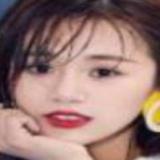}
          \end{subfigure}
      \end{subfigure}
      \hfill 
      \begin{subfigure}[b]{1\linewidth}
          \begin{subfigure}[b]{0.25\columnwidth}
            \centering
            \includegraphics[width=\linewidth]{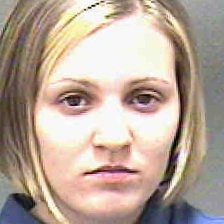}
          \end{subfigure}
          \hfill 
          \begin{subfigure}[b]{0.25\columnwidth}
            \centering
            \includegraphics[width=\linewidth]{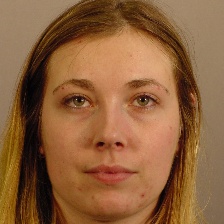}
          \end{subfigure}
          \hfill 
          \begin{subfigure}[b]{0.25\columnwidth}
            \centering
            \includegraphics[width=\linewidth]{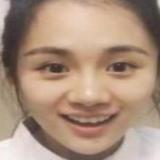}
          \end{subfigure}
      \end{subfigure}
  \end{subfigure}
  \caption{Examples of images of the same subject with (top) and without (bottom) makeup on the MORPH Caucasian (left), Notre Dame (middle), and AFD (right) datasets}
  \label{fig:makeup_examples}
  \vspace{-1em}
\end{figure}

Facial cosmetics is assumed to have two effects in face recognition: (a) different people wearing the same style of makeup are more likely to have a higher match score than two people without makeup; (b) images of the same person pair with and without makeup are likely to have a lower score than images where their makeup condition is the same.

\begin{figure*}[t]
  \begin{subfigure}[b]{1\linewidth}
      \begin{subfigure}[b]{0.24\linewidth}
          \begin{subfigure}[b]{1\columnwidth}
            \centering
            \includegraphics[width=\linewidth]{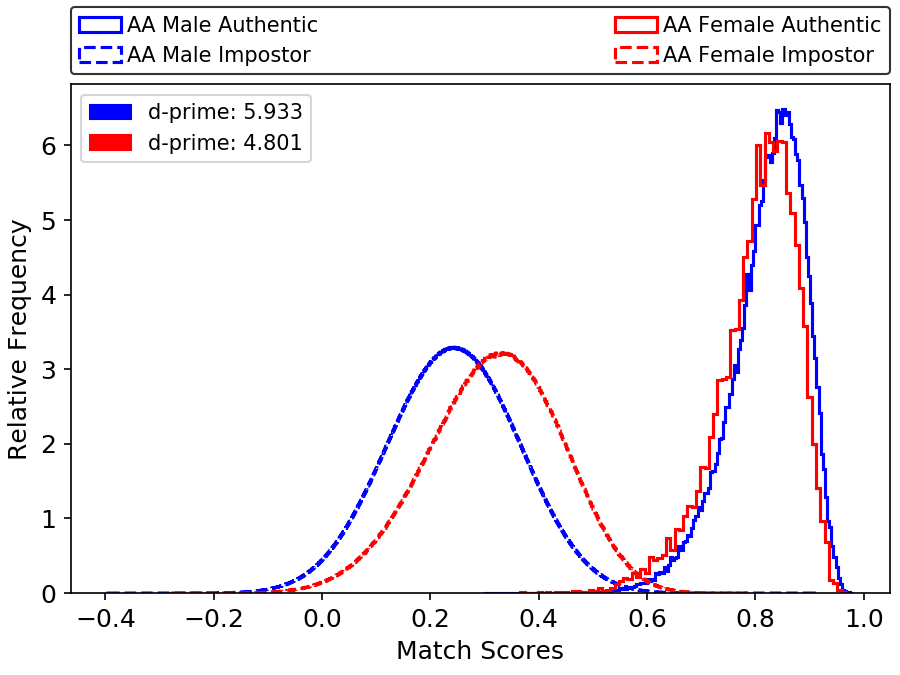}
          \end{subfigure}
      \end{subfigure}
      \hfill 
      \begin{subfigure}[b]{0.24\linewidth}
          \begin{subfigure}[b]{1\columnwidth}
            \centering
            \includegraphics[width=\linewidth]{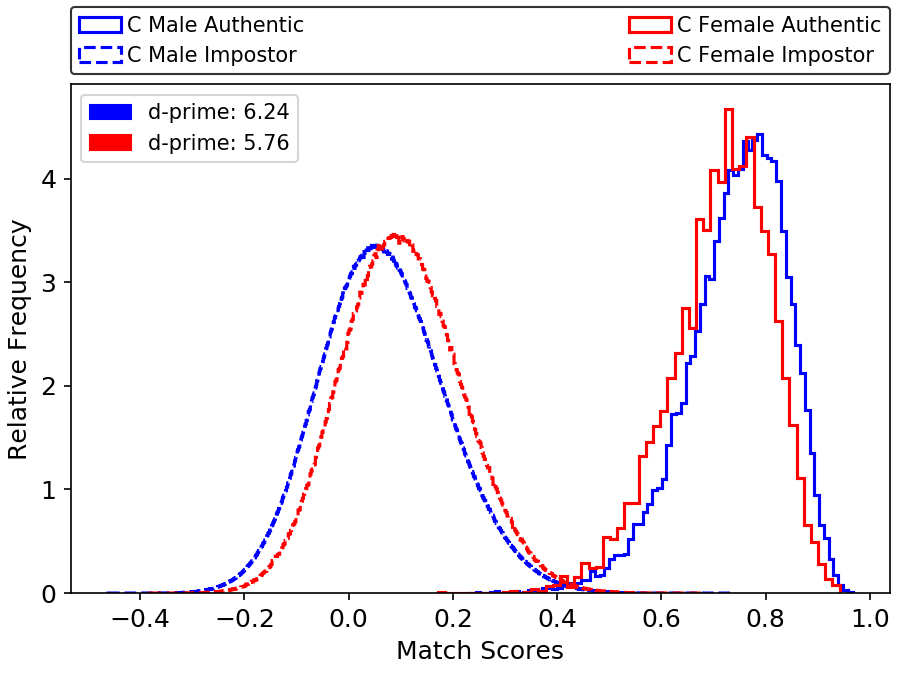}
          \end{subfigure}
      \end{subfigure}
      \hfill 
      \begin{subfigure}[b]{0.24\linewidth}
          \begin{subfigure}[b]{1\columnwidth}
            \centering
            \includegraphics[width=\linewidth]{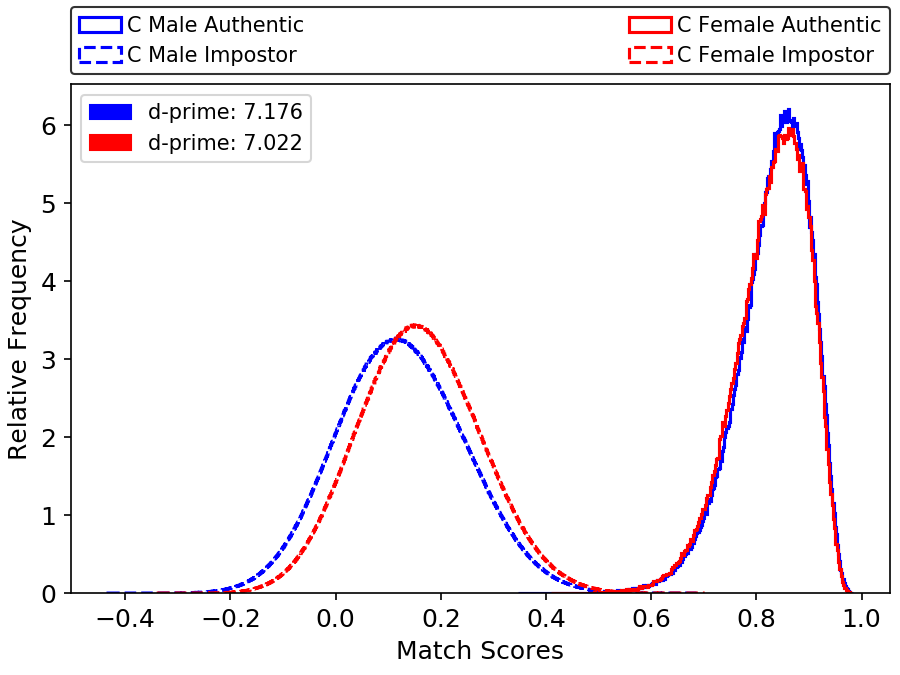}
          \end{subfigure}
      \end{subfigure}
      \hfill 
      \begin{subfigure}[b]{0.24\linewidth}
          \begin{subfigure}[b]{1\columnwidth}
            \centering
            \includegraphics[width=\linewidth]{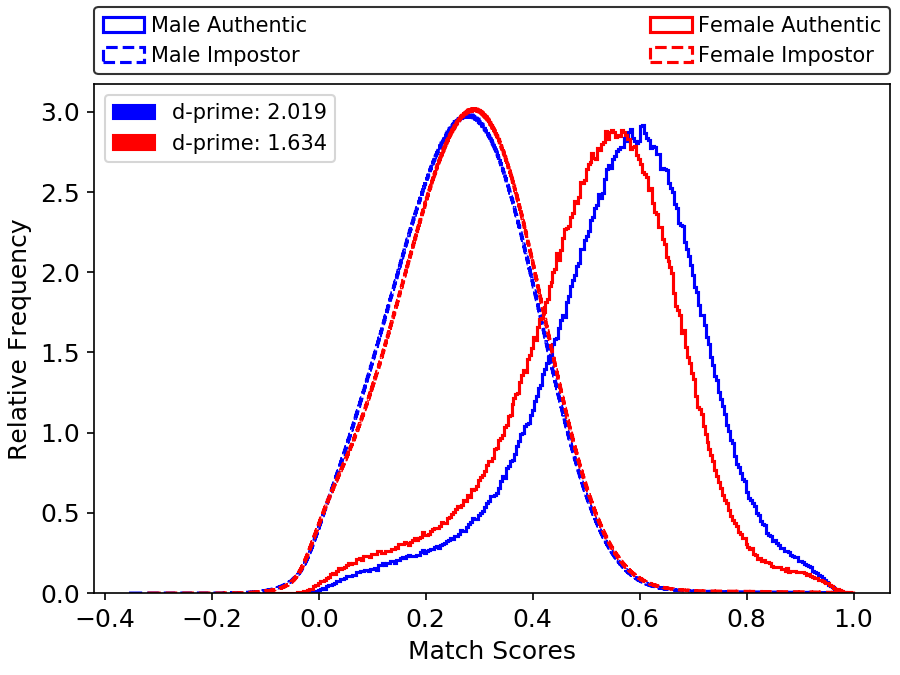}
          \end{subfigure}
      \end{subfigure}
  \end{subfigure}
  \begin{subfigure}[b]{1\linewidth}
      \begin{subfigure}[b]{0.24\linewidth}
          \begin{subfigure}[b]{1\columnwidth}
            \centering
            \includegraphics[width=\linewidth]{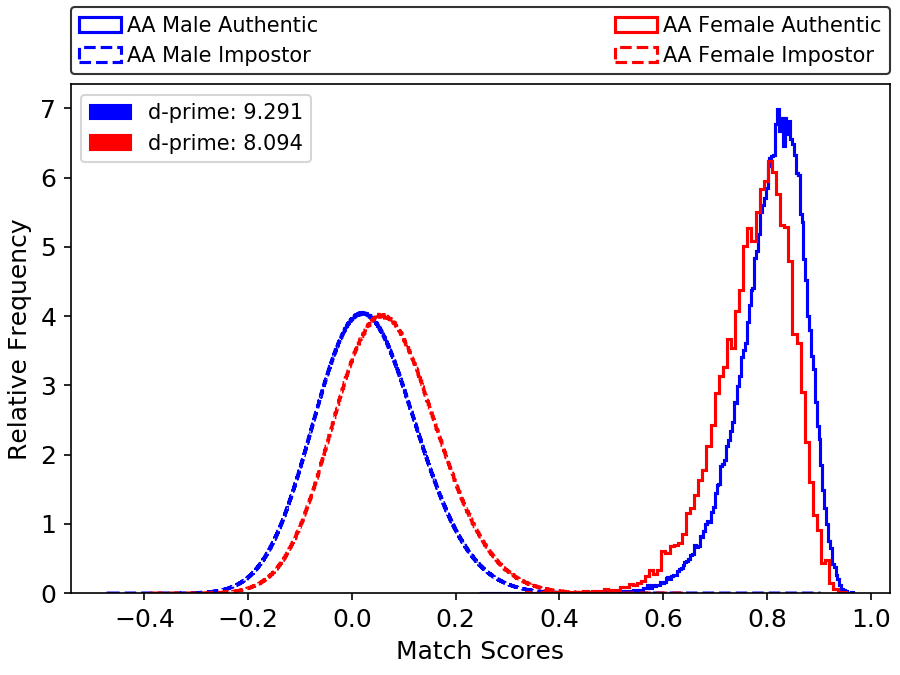}
          \end{subfigure}
          \caption{MORPH African American}
      \end{subfigure}
      \hfill 
      \begin{subfigure}[b]{0.24\linewidth}
          \begin{subfigure}[b]{1\columnwidth}
            \centering
            \includegraphics[width=\linewidth]{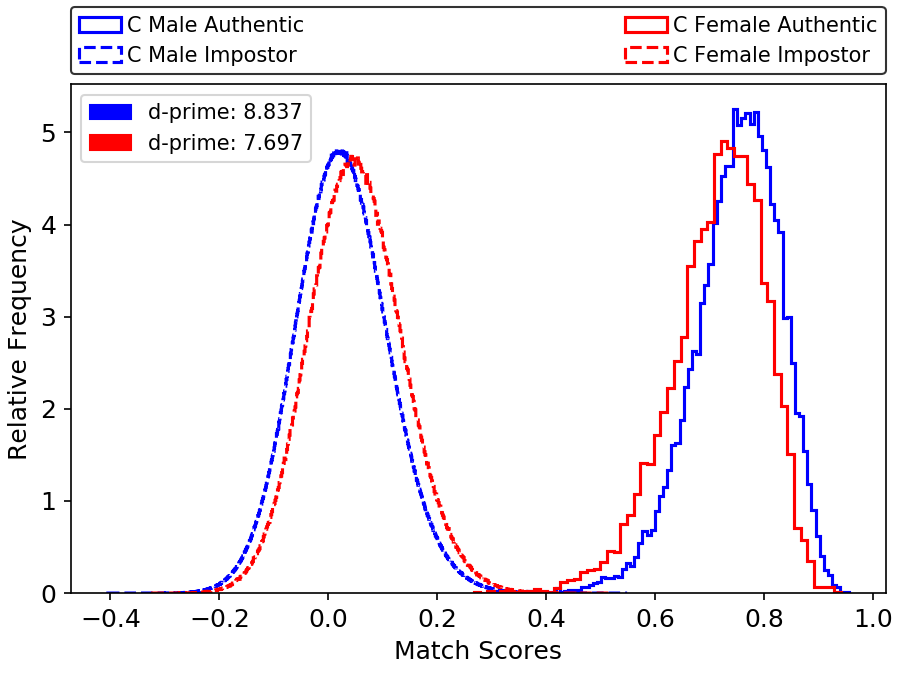}
          \end{subfigure}
          \caption{MORPH Caucasian}
      \end{subfigure}
      \hfill 
      \begin{subfigure}[b]{0.24\linewidth}
          \begin{subfigure}[b]{1\columnwidth}
            \centering
            \includegraphics[width=\linewidth]{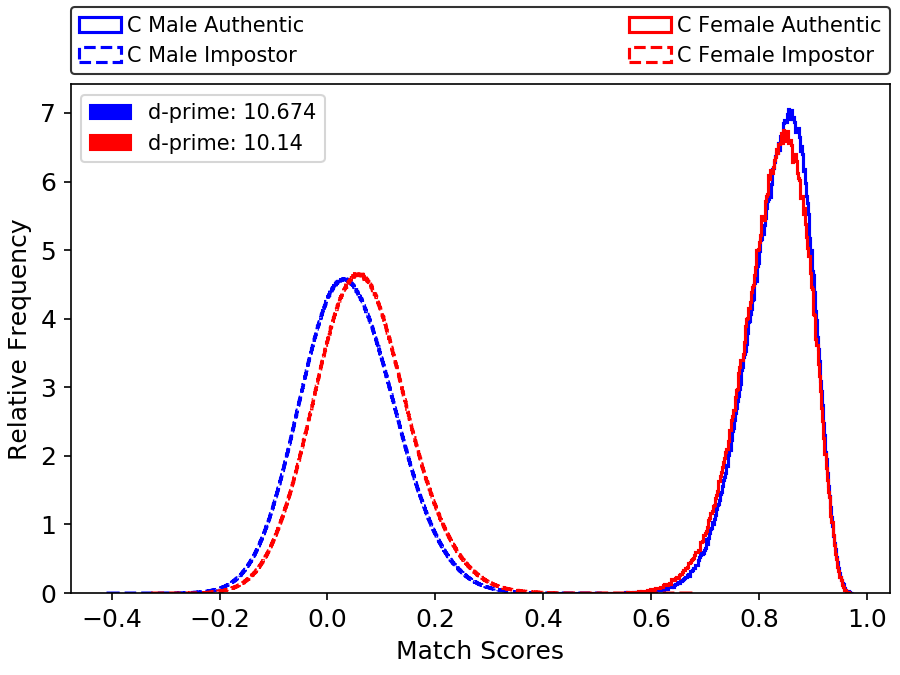}
          \end{subfigure}
          \caption{Notre Dame}
      \end{subfigure}
      \hfill 
      \begin{subfigure}[b]{0.24\linewidth}
          \begin{subfigure}[b]{1\columnwidth}
            \centering
            \includegraphics[width=\linewidth]{images/afd/hist_arcface_gender_afd.png}
          \end{subfigure}
          \caption{AFD}
      \end{subfigure}
  \end{subfigure}
  \caption{Male and female authentic and impostor distributions using a CNN trained with gender balanced VGGFace2 dataset (top) and gender balanced MS1MV2 dataset (bottom).}
  \label{fig:aut_imp_balanced}
  \vspace{-1em}
\end{figure*}

To detect facial cosmetics, we used the Microsoft Face API, which detects eye makeup and lip makeup separately.
In our experiments, we removed any image predicted to have either eye or lip makeup.
As the MORPH African-American female cohort had only 58 images predicted as having makeup, we did not experiment with this dataset.
After removal of makeup-predicted images, the female makeup-free data contains: 1,001 images of 394 females on the MORPH Caucasian dataset; 2,422 images of 130 females of the MORPH Caucasian dataset; and 1,051 images of 337 females of the AFD dataset.
Examples of subjects with and without makeup are shown in Figure \ref{fig:makeup_examples}.

Figure \ref{fig:aut_imp_makeup} shows the genuine and impostor distributions for all females and only females without makeup.
The effect (b) is supported by the genuine distribution shown, as the genuine distribution for females without makeup had a shift towards higher match scores.
However, the effect (a) is not seen here, in fact, the impostor distribution is showing the opposite: females without makeup have a higher impostor similarity score on average.

Table \ref{tab:d_prime} shows the comparison of females without makeup d-prime to males d-prime. For all datasets that females had makeup removed, the difference in d-prime between males and females decreased, achieving the lowest difference in d-prime between males and females.


\subsubsection{Balanced Training Dataset}
Perhaps the most commonly speculated cause for the lower accuracy on females is gender imbalance in the training dataset.
The ArcFace matcher was trained using the MS1MV2 dataset, which is only around 27\% female.
To test whether an explicitly balanced training set could eliminate the accuracy difference, we trained two separate ResNet-50 \cite{resnet} networks with combined margin loss (which combines CosFace \cite{cosface}, SphereFace \cite{sphereface} and ArcFace \cite{arcface} margins) using a subsets of the VGGFace2 dataset \cite{vggface2} and the MS1MV2 dataset, that we balanced to have the exactly same number of male and female images and subjects. 

The male and female genuine and impostor distributions using the two gender balanced training sets are shown in Figure \ref{fig:aut_imp_balanced}.
The model trained with the gender balanced MS1MV2 achieved better distributions for all datasets and both genders compared to the model trained with the gender balanced VGGFace2.
For Notre Dame, on both models, the d-prime difference between males and females is smaller than before.
However, both of the explicitly-balanced training datasets still result in the female impostor and genuine distributions being closer together than the male distributions. 

\section{Conclusions and Discussion}

Our results show that although ROC curves may appear similar for women and men in constrained datasets, they in fact experience very different face recognition accuracy.
Women and men achieve the same FMR at very different thresholds.
The cause for the higher thresholds used for females is the impostor distribution being shifted towards higher values.
Along with it, the females genuine distribution is also shifted, but towards lower values.
Thus women are disadvantaged relative to men both on FMR and FNMR.

We showed that although women have many more images with non-neutral facial expression,
restricting the images to a neutral-expression-only subset does not affect the pattern of genuine and impostor distributions being closer together than for men.
Moreover, when removing images with pitch angle higher than 5 degrees, the same phenomenon is still present.
Also, when removing any image with forehead occluded by hair, the pattern persists.

Our results show that when only makeup-free images are used, the female genuine distribution moves towards higher values.
This presumably results from eliminating matches across images with and without makeup.
However, the impostor distribution also moves towards higher values.
Thus going to only no-makeup images for females makes the initial FNMR better but also makes the initial FMR worse. 

Selecting only images that satisfies all the factors examined together greatly reduces the amount of data.
The dataset with the most female data after removing all the conditions is the MORPH African-American dataset.
However, the female genuine/impostor shift still exists, with d-prime of 10.96 and 9.48 for male and female, respectively.
Further research into selecting multiple conditions requires a larger or a different dataset.

As deep learning models are driven by data, gender imbalance in training data could be the reason for lower female accuracy. Our results with trained models on two gender-balanced datasets suggests that training data is not the main reason for lower female accuracy, as in general, the same genuine/impostor distribution pattern persists.

This study has found ``negative results'' for each of the speculated causes of gender accuracy difference that we have examined.
This motivates looking for a cause that is more intrinsic to the difference(s) between males and females.
One line of future research is to look at face morphology differences between individual men and women and between men and women as groups.
Anecdotal supporting evidence to look for a cause in this area can be found in recent work by Muthukumar \cite{Muthukumar}.

{\small
\bibliographystyle{ieee}
\bibliography{refs}
}

\end{document}